\documentclass[final]{cvpr}

\usepackage{times}
\usepackage{epsfig}
\usepackage{graphicx}
\usepackage{amsmath}
\usepackage{amssymb}
\usepackage{subfigure}
\usepackage{multirow}
\usepackage{multicol}
\usepackage{booktabs}

\usepackage[pagebackref=true,breaklinks=true,colorlinks,bookmarks=false]{hyperref}
\usepackage{cleveref}

\def\cvprPaperID{4560} 
\def\confYear{CVPR 2021}

\begin{document}

\title{Few-Shot Human Motion Transfer by \\Personalized Geometry and Texture Modeling}

\author{Zhichao Huang$^\text{1}$, Xintong Han$^\text{2}$, Jia Xu$^\text{2}$, Tong Zhang$^\text{1}$\\
$^1$The Hong Kong University of Science and Technology\qquad$^2$Huya Inc\\
{\tt\small zhuangbx@connect.ust.hk, \{hanxintong, xujia\}@huya.com, tongzhang@ust.hk}
}

\maketitle
\pagestyle{empty}  
\thispagestyle{empty}
\begin{abstract}
  We present a new method for few-shot human motion transfer that achieves realistic human image generation with only a small number of appearance inputs. Despite recent advances in single person motion transfer, prior methods often require a large number of training images and take long training time. One promising direction is to perform few-shot human motion transfer, which only needs a few of source images for appearance transfer. However, it is particularly challenging to obtain satisfactory transfer results. In this paper, we address this issue by rendering a human texture map to a surface geometry (represented as a UV map), which is personalized to the source person. Our geometry generator combines the shape information from source images, and the pose information from 2D keypoints to synthesize the personalized UV map. A texture generator then generates the texture map conditioned on the texture of source images to fill out invisible parts. Furthermore, we may fine-tune the texture map on the manifold of the texture generator from a few source images at the test time, which improves the quality of the texture map without over-fitting or artifacts. Extensive experiments show the proposed method outperforms state-of-the-art methods both qualitatively and quantitatively. Our code is available at \url{https://github.com/HuangZhiChao95/FewShotMotionTransfer}.
\end{abstract}

\section{Introduction}

Human motion transfer \cite{everybodydancenow,softgatedpose,clothflow,fusionimage,densefkiwpose,disentangledpose,deformablegan,dancedance, zhu2019progressive} 
generates videos of one person that takes the same motion as the person in a target video, which has huge potential applications in virtual characters, movie making, \etc.
The rapid growth of generative networks \cite{gan} and image translation frameworks \cite{pix2pix, pix2pixhd} enables generating photo-realistic images for human motion transfer. Basically, one would extract the pose sequence of the target video and take the pose as the input to generate the video for a new person. 

\begin{figure}[!t]
   \centering
   \includegraphics[width=\linewidth]{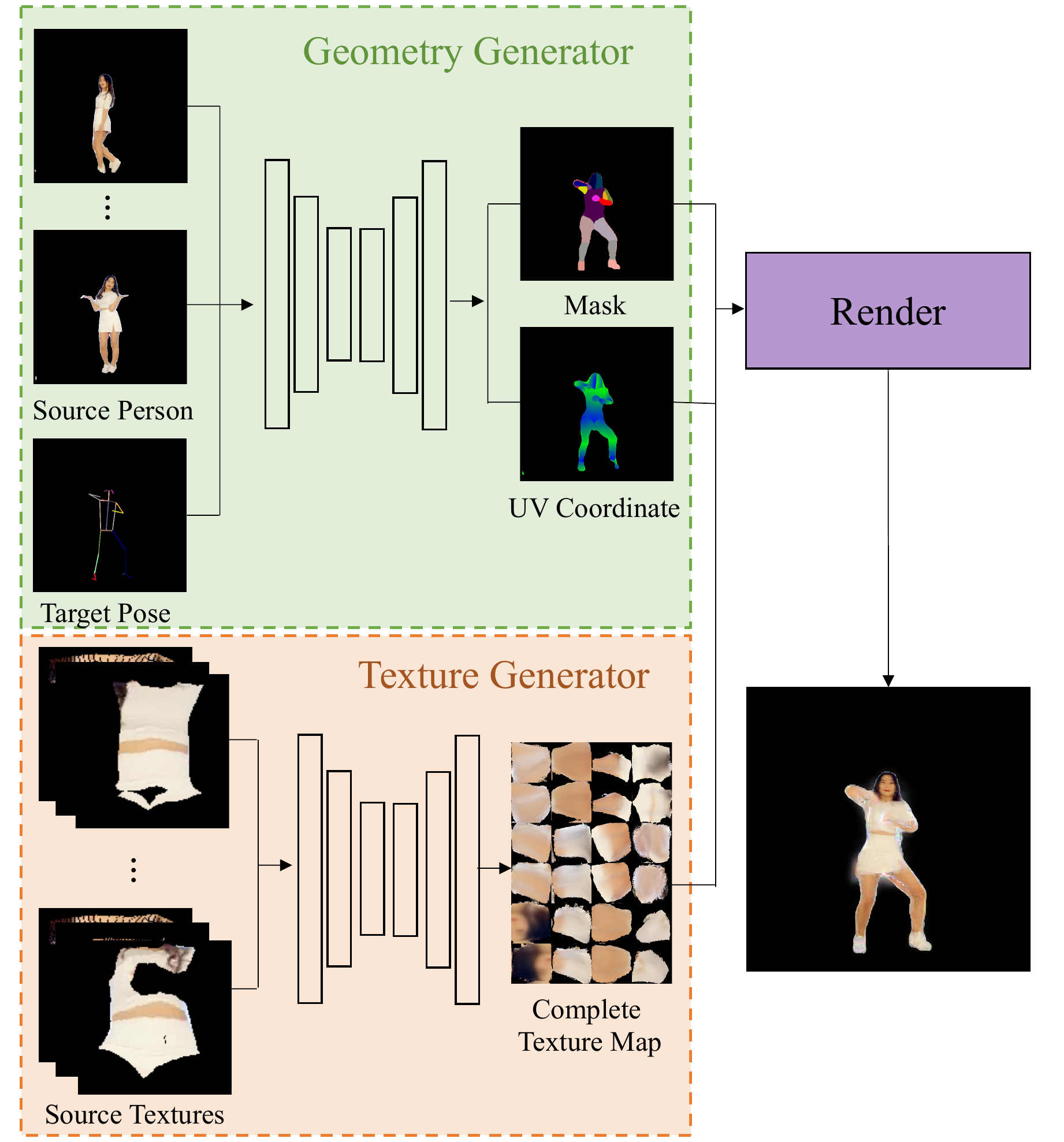}
   \caption{Method overview. Our method contains three key components: a geometry generator for generation of a personalized UV map, a texture generator to fill out incomplete texture, and a neural renderer rendering the human image. Detailed network structures are illustrated in \Cref{fig:detail_diagram}.}
   \label{fig:diagram}

\end{figure}

The appearance of the new person can be provided in two ways. One type of models aim to train an individual model for a specific person. To obtain such a model, one needs to collect a large number of images for the new person and trains a network to translate the pose to the image of the person \cite{everybodydancenow, vid2vid}. Then the appearance information is stored in the weights of the network and the image of new pose can be directly generated by taking the new pose as input. 
However, such methods need a large amount of training data and training time to obtain a model for the new person, which hinders the applications of these approaches.

For the other type, the information of appearance is provided by taking a few images of the new person as input. Few-shot human motion transfer requires the network to learn the complicated relationship between human appearance and pose by only looking at few human images. The relationship is very hard to learn and generalize to unseen people. Therefore, directly conditioning the output image on the pose and the appearance leads to poor quality. Some approaches warp the input appearance to the output with optical flows  \cite{fewshotvid} 
or affine transformations \cite{posewarp,dancedance} to generate a coarse pose of the new person. However, the mapping is usually inaccurate and fails to recover realistic human from the intermediate warping result. Even if the architecture gets more and more complicated, there are still many artifacts in the images of few-shot human motion transfer.


%

The appearance of one person at different pose is the same. Therefore, we can directly transfer the pixels from the source pose to the target pose without generating the pixels. DensePose \cite{densepose} provides the UV map of one person so that the texture can be transferred between different poses to synthesize the human at the new pose. However, directly using the original model to transfer the texture fails to generate realistic human image \cite{densepose, denseposetransfer}. 


The DensePose can be trained to better fit the generation of human and achieves high quality avatars \cite{texturedavatars}. However, their method is only suitable for single person and cannot be directly used for few-shot synthesis. On the one hand, their generator is not able to synthesize accurate geometry (\ie IUV representation) of different people whose shapes are different. On the other hand, their method cannot produce complete texture map from only few the source human images. We propose a new method that generalizes the algorithm for the few-shot scenario and get better results than previous few-shot approaches. As shown in \Cref{fig:diagram}, we use a geometry generator to generate personalized UV map given a target pose and a few source images. Meanwhile, a texture generator merges each incomplete texture map and hallucinates the invisible. Then the texture map is rendered to the UV map to generate an image with target pose and source appearance. The decoupling generation of personalized geometry and texture leads to better quality of motion transfer.

We summarize our contributions as follows:



1. We propose a geometry generator to predict an accurate personalized UV map and a texture generator to generate a complete texture map. These two generators work collaboratively for rendering high quality human images.

2. By training on multiple videos of multiple persons and fine-tuning on a few examples of an unseen person, our method successfully transfers geometry and texture knowledge to the new person.

3. Experiments demonstrate that our method generates better human motion transfer results than state-of-the-art methods both qualitatively and quantitatively.


\section{Related Work}

\noindent\textbf{Human Motion Transfer.} There have been a lines of work about synthesizing a human image in an unseen pose \cite{posewarp,everybodydancenow,esser2018variational,liu2019neural,denseposetransfer,si2018multistage}. Most of the methods implement a generator condition on 2D keypoints (or connection of the keypoints) of the pose. One type of the methods train different models for different persons. EDN \cite{everybodydancenow} utilize pix2pixHD \cite{pix2pixhd} framework to translate 2D skeleton to the image of a specific person. Vid2Vid \cite{vid2vid} uses more complicated network for generation of the video, which contains foreground-background separation and optical flow warping module. While single person model generates photo-realistic picture of the human, it requires collecting training data for each person and takes long time to train. 

Another type of the methods train single model to transfer motion for all persons. As appearance information needs to be obtained from the source images, which makes the task much more complicated, many papers add additional modules for synthesizing intermediate images and use them as one input for later generative networks. The additional modules include affine transformation \cite{posewarp, dancedance}, flow warping \cite{fewshotvid}, DensePose transfer \cite{denseposetransfer} and SMPL transfer \cite{densefkiwpose,liu2019liquid}. Our method also uses DensePose for modeling the geometry of the pose, but we directly render the texture to the DensePose to produce final outputs instead of using it as the intermediate layer. While our method requires accurate personalized UV map and high quality texture map, we omit directly generating the pixel. Multi-stage methods also generate the coarse intermediate images with neural network generators \cite{poseguided,si2018multistage,zhu2019progressive}. In addition, Siarohin \etal \cite{deformablegan} and Liu \etal \cite{liu2019liquid} modify blocks of the network to adapt the task. MonkeyNet \cite{monkeynet} does not depend on the 2D pose. Instead, it extracts and maps the keypoints of the source and target image and animates any objects by the motion of these keypoints. Moreover, fine-tuning on a few source images of one person can improve the quality of the output \cite{metapix}. Fine-tuning is also part of our method. However, we mainly fine-tune the texture map, which seldom overfits to the few source images at test time.

\noindent\textbf{Human Avatars.} 
Full-body human avatars are usually represented by textured animatable 3D human models. There have a large group of works on building 3D model from single-view or multi-view images \cite{ Alldieck_2019_CVPR, MGN, arch, lazova2019360, smpl}. Many 3D models of human are based on SMPL \cite{smpl}, a parametrized model describing the shape and pose of human. Lazova \etal \cite{lazova2019360} build fully-textured avatars from a single input image. Starting from SMPL, it uses neural network to model the displacement of geometry and complete partial texture of human. PIFu \cite{pifu3d} learns an implicit function to align surface and texture so that human avatar can be reconstructed from single-view or multi-view images. Aliaksandra \etal \cite{texturedavatars} learn the translation from skeleton to UV map from a video. And full texture is generated by directly optimizing the output image of neural rendering. Our method is close related to \cite{texturedavatars}, but we need to train one model for all persons in a few-shot setting instead of training a single model for each person. So our model should be able to synthesize personalized geometry and hallucinate unseen texture map.


\section{Personalized Geometry and Texture Model}

Given a target pose $P$ and a few images of the source person $I_{1}, ..., I_{b}$, our goal is to learn a model $f$ that synthesizes an image 
\begin{equation}
   \widehat I = f(P, \{I_j\}_{j=1}^b),
\end{equation}
in which the generated person has the pose $P$ and the appearance of source images $I_{1}, ..., I_{b}$. In this paper, we assume the source person images share a fixed background, so we can separate the generation of human image $\widetilde I$ and background $B$ as:
\begin{equation}
  \widehat I = (1-m)\odot B + m\odot \widetilde{I},
  \label{eqn:fgbg}
\end{equation}
where $m$ is the soft mask indicating the human image foreground $\widetilde{I}$, and $\odot$ represents element-wise product. The mask $m$ can be easily obtained with an off-the-shelf person segmentation model like DeepLab V3 \cite{humanseg}.

\subsection{Neural Human Rendering} \label{sec:rendering}

$\widetilde{I}$ can be directly generated from the pose $P$ with an image-to-image translation network \cite{posewarp,everybodydancenow,pix2pix}. However, these methods fail to model complex geometric changes and detailed textures of the person, resulting in low-quality outputs especially when only few training samples of the source person are available. To mitigate this issue, we takes a neural rendering based approach \cite{texturedavatars,thies2019deferred} by dividing the image synthesis into two parts: generation of a personalized human geometry (UV map) and generation of personalized human texture (texture map). And the person image can be generated by sampling the texture map according to the UV map as shown in Figure \ref{fig:diagram}. On the one hand, while human geometry varies across different persons, the variation is much smaller than that of person images. Therefore, generating personalized geometry is easier than directly generating the human image. Texture map, on the other hand, is fixed for one person without complex geometric changes. So it can be effectively learned from the source textures.

DensePose \cite{densepose} descriptors have been widely adopted for disentangling the generation of human geometry and texture \cite{denseposetransfer,texturedavatars}. We follow this line of work and subdivide human's body into $n=24$ non-overlapping parts. The $k$-th body part  ($k = 1,2...,n$) is parameterized by a 2D coordinate $(C^{2k}, C^{2k+1})$ (\ie, a UV map) and associated with a texture $T^k$ as shown in Figure \ref{fig:diagram}. Then, we can render the $k$-th part with bilinear interpolation: 
\begin{equation}
   R^k[x,y] = T^k[C^{2k}[x,y], C^{2k+1}[x,y]]. \label{eqn:part_render}
\end{equation}
For pixel $[x,y]$ of the image, we assign a score $S^k[x,y]$ to the $k$-th part of the DensePose, representing the probability that the pixel belongs the $k$-th part. $\sum_{k=1}^{n+1}S^k[x, y] = 1$, where $S^{n+1}[x,y]$ indicates the probability that $[x,y]$ belongs to the background (\ie, $(1-m)$ in \Cref{eqn:fgbg}). By summing up the rendered part weighted by its probability, we generate the image of the human as:
\begin{equation}
   \widetilde{I} = \sum_{k=1}^n S^k[x,y] R^k[x,y]. \label{eqn:render}
\end{equation} 

In this work, we design a geometry generator (\Cref{sec:geometry}) to estimate body part score $S^k$ and its UV map $(C^{2k}, C^{2k+1})$, as well as a texture generator (\Cref{sec:texture}) for generating body part texture $T^k$. These two generators can be trained collaboratively and render the human image with transferred motion using \Cref{eqn:part_render,eqn:render}.

\begin{figure*}[t!]
   \centering
   \subfigure[Geometry Generator (\Cref{sec:geometry})]{
       \begin{minipage}[t]{0.59\linewidth}
           \centering
           \includegraphics[height=8cm]{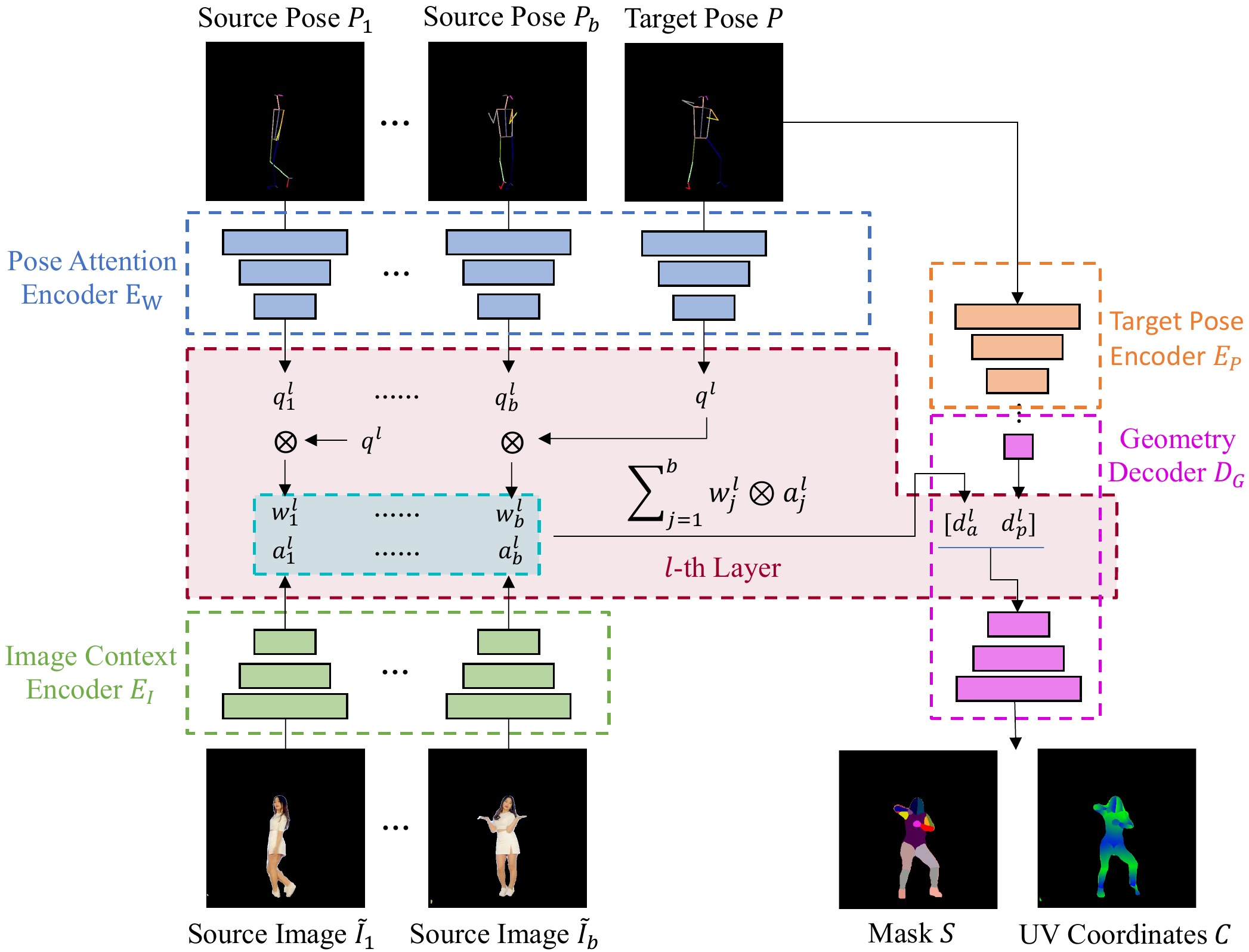}
       \end{minipage}
       \label{fig:coordinate_diagram}

   }
   \subfigure[Texture Generator (\Cref{sec:texture})]{
       \begin{minipage}[t]{0.36\linewidth}
           \centering
           \includegraphics[height=8cm]{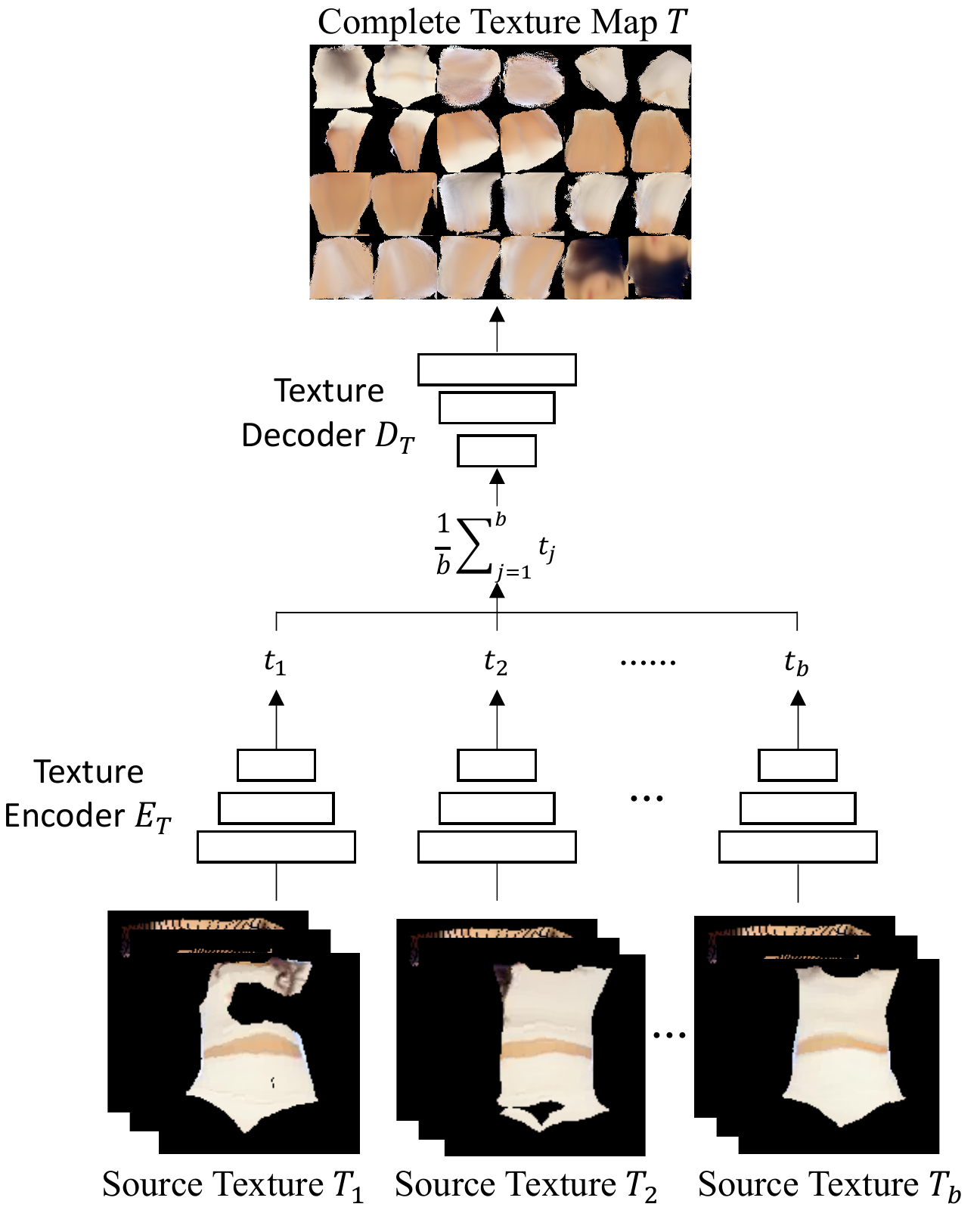}

       \end{minipage}
       \label{fig:texture_diagram}

   }
   \caption{Architecture of our proposed geometry generator (\Cref{sec:geometry}) and texture generator (\Cref{sec:texture}). The generated geometry and texture are finally used to render a reconstructed human image (\Cref{sec:rendering}).}
   \vspace{-0.5cm}
   
\label{fig:detail_diagram}
\end{figure*}

\subsection{Geometry Generator} \label{sec:geometry}


As shown in \Cref{fig:coordinate_diagram}, the body UV geometry not only depends on the target pose $P$ but also needs to be personalized, varying across different persons. To this end, our geometry generator $G_\phi$ takes the pose $P$ as input to generate the geometry with desired pose, and at the same time, it also takes the source human images $\widetilde{I}_{1},...,\widetilde{I}_{b}$ to model personalized details (\eg, hair style, clothing, body shapes):
\begin{align}
   C &= G_\phi^C(P,\{\widetilde{I}_{1}, \widetilde{I}_{2}, ..., \widetilde{I}_{b}\}), \nonumber\\
   S &= G_\phi^S(P,\{\widetilde{I}_{1}, \widetilde{I}_{2}, ..., \widetilde{I}_{b}\}), 
   \label{eqn:coordinate_generator}
\end{align}
where $b$ is the number of source human images provided as input to the geometry generator. We denote $C^k$ for the $k$-th channel of $C \in \mathbb{R}^{48\times H \times W}$, and $C^{2k}, C^{2k+1}$ are the U- and V-coordinate for the $k$-th body part, respectively. $S \in \mathbb{R}^{25\times H \times W}$, whose $k$-th channel $S^k$ is the soft assignment mask for $k$-th part of the DensePose, represents the probability that a pixel belongs to which part of the body. 
Note that during training, $P$ is different from the poses of $\widetilde{I}_{1}, ..., \widetilde{I}_{b}$, so that the network is forced to rely the on the target pose $P$ when extracting geometric information.

As shown in \Cref{fig:coordinate_diagram}, the geometry generator contains three encoders and one decoder: an image context encoder $E_I$, a pose attention encoder $E_W$, a target pose encoder $E_P$, and a geometry decoder $D_G$. The target pose encoder $E_P$ extracts the information of target pose $P$ as input to the geometry decoder $D_G$ to ensure the generated geometry reflects the desired pose. The image context encoder $E_I$ extracts personalized body information from $\widetilde{I}_{1},..., \widetilde{I}_{b}$. The pose attention encoder $E_W$ then compares the similarity between target pose $P$ and source poses  ${P}_{1}, ..., {P}_{b}$ to determine which source images should be paid more attention to for incorporating personalized shape details in the generated geometry output by $D_G$. 


We implement our geometry generator with a U-Net \cite{unet} like architecture. And we perform the above feature interaction at each corresponding layer of the three encoders and the decoder (\ie, layers with the same spatial resolution). Consequently, the architecture captures multi-level features from the input images and poses at different resolutions, which helps to generate high quality geometry with accurate personalized shape and pose. 

More specifically, the encoder $E_I$, $E_W$, $E_P$ and decoder $D_G$ have the same number of layers $L$. We define $E^l$ as $l$-th layer of the encoder, and $D^l$ as $(L-l)$-th layer of the decoder. At $l$-th layer of the encoders, we calculate the feature for $(l+1)$-th layer as:
\begin{equation}
   a_{j}^{l+1} = E_I^l(a_{j}^l),\quad q_{j}^{l+1} = E_W^l(q_{j}^l),
   \label{eqn:encoder}
\end{equation}
\begin{equation}
   w_{j}^l = (q_{j}^l)^\top \otimes q^l, \quad d_a^l = \textstyle \sum_k w_{j}^l \otimes a_{j}^l,
   \label{eqn:attention}
\end{equation}
\begin{equation}
   d_p^{l+1} = D_G^l([d_a^l, d_p^l]),
   \label{eqn:decoder}
\end{equation}
where $j = 1, ..., b$ is the index of the source images, $a_{j}^0 = \widetilde{I}_{j}$, $q_{j}^0 = P_{j}$, and $d_p^0 = E_P(P)$. $\otimes$ denotes matrix multiplication. In \Cref{eqn:encoder}, $E_I^l$ and $E_W^l$ encode features from previous layer. \Cref{eqn:attention} shows how we merge the personalized shape information from different source images using an attention mechanism. As different source images carry different shape information, when predicting the geometry $C$ and $S$, we give different weights to these inputs according to their similarities with the target pose $P$. For instance, if $P$ describes the side view of the person, it may be hard to infer the detailed shape geometry from front-side images. So, we should give more attention to the images whose pose is similar to the target pose. In \Cref{eqn:attention}, we reshape $a_{j}^l$ and $q_{j}^l$ into $\mathbb{R}^{C_l \times N_l}$ where $C_l$ is the number of channels and $N_l = H_l \times W_l$ denotes the spatial size of the feature map, then we perform matrix product with $\otimes$. At last, as shown in \Cref{eqn:decoder}, $d_a^l$ is reshaped back into $C_l \times H_l \times W_l$, concatenated with $d_p^l$, and fed into $D_G^l$.

\subsection{Texture Generator} \label{sec:texture}
Our texture generator is responsible for generating a full human texture map $T$ given the source images. An intuitive approach would be directly extracting a DensePose texture map from each source image ${I}_{j}$ and aggregating them (\eg, through spatial average or max pooling) to get a merged texture map. However, such merged texture map is usually incomplete since not all body parts are visible from the given source images. Plus, due to inaccurate DensePose estimation, this texture map is unrealistic and lacks of fine texture details.  To solve this problem, we introduce a texture generator $H_\theta$ that takes the textures ${T}_{1}, ...,  {T}_{b}$ extracted by DensePose from source images  ${I}_{1}, ...,  {I}_{b}$ to synthesize the complete texture $T$ in a learnable fashion:
\begin{equation}
   T = H_\theta({T}_{1}, {T}_{2}, ..., {T}_{b}).
\end{equation}
The architecture of our texture generator is a vanilla encoder-decoder as shown in \Cref{fig:texture_diagram}. We reshape the input texture ${T}_{j}$ from $24\times 3\times H_T \times W_T$ to $72 \times H_T \times W_T$ and feed it to $H_\theta$. The encoder $E_T$ encodes each texture $T_{j}$ to the bottleneck embedding denoted as $t_{j}$. We merge the textures of different source images by taking average of their embedding features: 
\begin{equation}
   t = \frac{1}{b}\sum_{j=1}^b t_{j}.
   \label{eqn:embed}
\end{equation}
Note that the architecture allows different numbers of input textures. And $t$ is then fed into decoder $D_T$ to produce the complete texture map of size ${72\times H_T\times W_T}$, which is finally reshaped into $T \in \mathbb{R} ^{24\times 3\times H_T \times W_T}$. Our texture generator not only produces the complete texture map $T$, but also defines a manifold of human body textures. As we will discuss later, we may fine-tune the embedding $t$ at test-time to improve the quality of texture map $T$.



\section{Training}
Our training process consists of three stages: an initialization stage, a multi-video training stage, and an optional few-shot fine-tuning stage.

\subsection{Initialization}

Training our geometry and texture generators with image-level reconstruction loss from scratch is infeasible, as the model does not have any prior information about human body. For example, it is impossible for the geometry generator to learn body semantics and output a human mask $S$, of which each channel corresponds to a specific body part, without any explicit supervision. Thus, we follow \cite{texturedavatars} and use the output of an off-the-shelf DensePose extractor \cite{densepose} to initialize our networks.



\noindent\textbf{Geometry Generator.} For a target ground truth image $I$ we aim to reconstruct, we take the pseudo ground truth body part mask $S^*$ and UV-coordinate $C^*$ extracted by the DensePose model as the supervision signal to initialize the geometry generator $G_\phi$ by minimizing:
\begin{align}
   L_{C} = &\sum_{k=1}^{24}(\|S^{*k} \odot (C^{2k}-C^{*2k})\|_1+ \nonumber\\& \|S^{*k}\odot (C^{2k+1}-C^{*2k+1})\|_1), \\
   L_S = &L_{CE}(S, S^*),
\end{align}
where $C$ and $S$ are the outputs of our geometry generator as in \Cref{eqn:coordinate_generator}. $L_{C}$ is the L1 norm between $C^*$ and $C$ on the given body part. And $L_S$ is the cross-entropy loss as used in semantic segmentation that guides the generator to predict the same body part masks as $S^*$.

\noindent\textbf{Texture Generator.} The texture generator is initialized by requiring its output to have the same texture on the visible part of its inputs. Suppose $\sigma_{j}$ is the binary mask indicating the visible part of the input texture $T_{j}$, we optimize $H_\theta$ with the following pixel L1 loss:
\begin{equation}
   L_{T} = \sum_{j=1}^b\| \sigma_{j}\odot (T - T_{j})\|_1
\end{equation}

\subsection{Multi-video Training}
After the initialization, our model can roughly generate human geometry and texture. However, the generated geometry is similar to the pre-trained DensePose outputs that lack personalized shape details. Also, the texture map is only coarsely rendered, missing important detailed textures that are necessary for generating realistic humans.

To this end, we train our generators on multiple training videos after the initialization. In each training mini-batch, we only sample data from one person so that the texture $T$ and feature $d_a$ can be shared across different target pose $P$. Besides, training time is saved as $T$ and $d_a$ only need to be generated once for all target pose in this mini-batch. We consider the following losses during training.


\noindent\textbf{Image Loss.} Image loss makes the reconstructed human $\widetilde{I}$ to be closer the ground truth human image $m\odot I^*$ in both image space and the feature space of a neural network \cite{perceptual}. Suppose $\Phi^v$ is an intermediate feature of a pre-trained VGG-19 network \cite{vgg} at different layers. The image loss is:
\begin{equation}
   L_I = \|\widetilde{I} - m\odot I^*\|_1 + \sum_{v=1}^N \|\Phi^v (\widetilde{I}) - \Phi^v (m\odot I^*)\|_1,
\end{equation}

\noindent\textbf{Mask Loss.} 
The mask loss is a cross-entropy (CE) loss between the generated background mask $S^{25}$ and a pseudo ground truth background mask $1 - m$ output by a SOTA segmentation model \cite{humanseg}:
\begin{equation}
   L_M = L_{\text{CE}}(S^{25}, 1-m).
\end{equation}

\begin{figure}[t]
    \centering
    \subfigure[With Regularization]{
        \begin{minipage}[t]{0.35\linewidth}
            \centering
            \includegraphics[width=\linewidth]{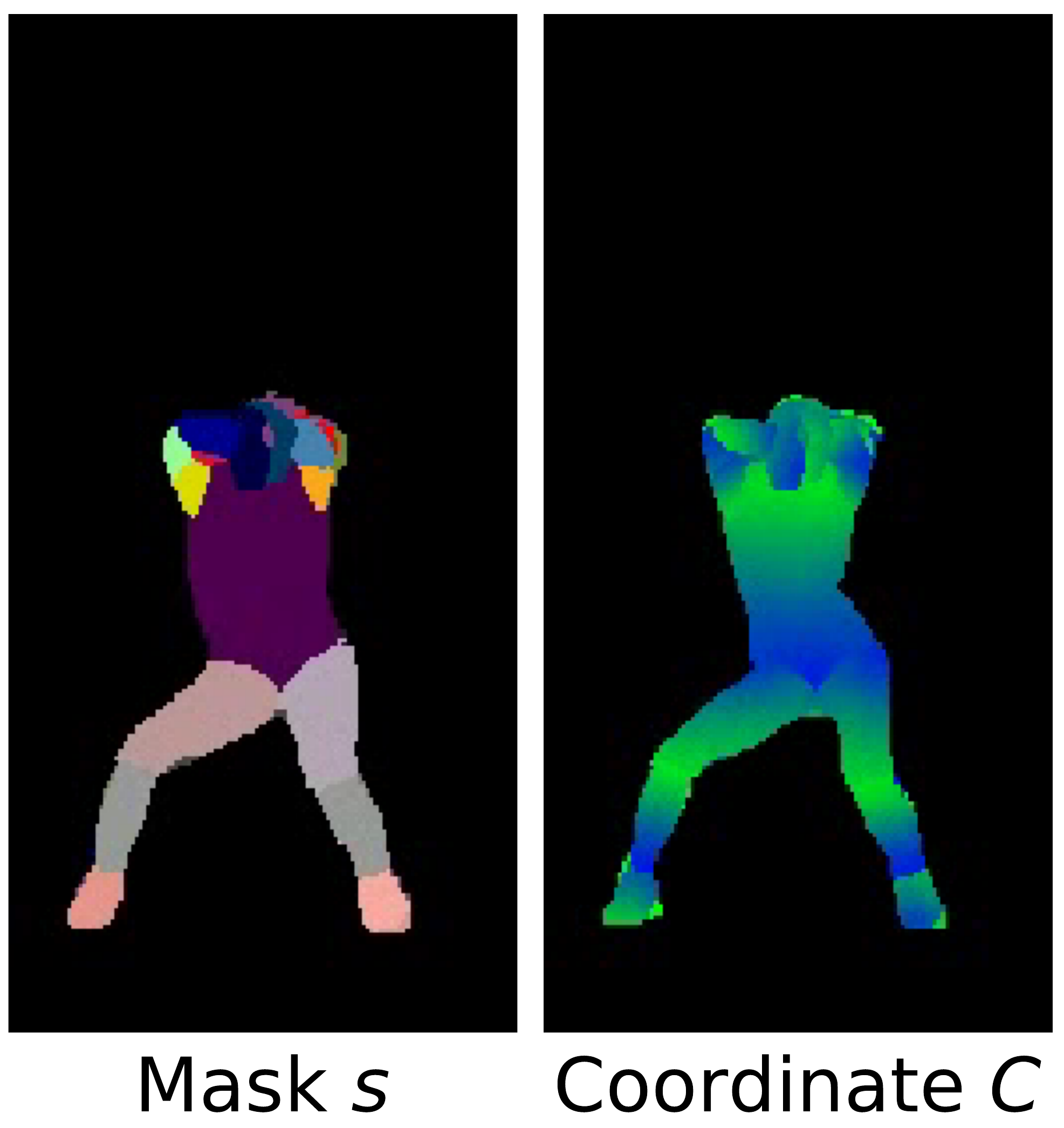}    
        \end{minipage}
    }
    \qquad
    \subfigure[Without Regularization]{
        \begin{minipage}[t]{0.35\linewidth}
            \centering
            \includegraphics[width=\linewidth]{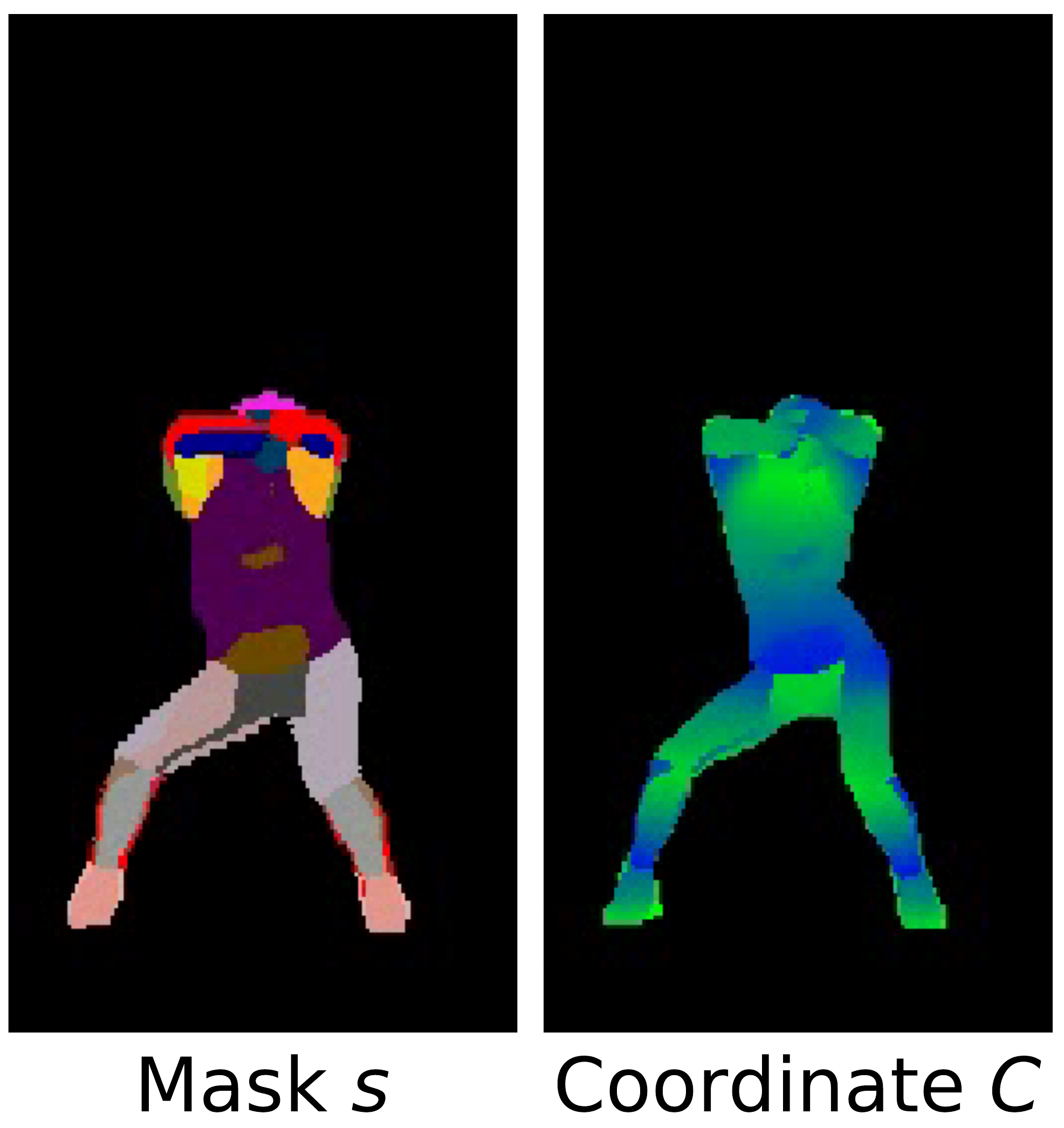}   
        \end{minipage}
    }
    \caption{Impact of adding regularization terms (\Cref{eqn:lrc,eqn:lrt}) to the loss. The belly and leg regions are messy without regularization.}
    \label{fig:constraint}
    \vspace{-0.4cm}
\end{figure}


\noindent\textbf{Generator Regularization Loss.} The rendering process makes the optimization of geometry and texture ambiguous. There would be infinite combinations of geometry and texture map to render the same human image. As the neural network have high flexibility, we need to constrain dramatic changes of texture, coordinates and mask. Otherwise, the geometry and texture generators are prone to overfitting and cannot generalize to unseen people and poses. As shown in \Cref{fig:constraint}, the mask and coordinate gets irregular without the regularization. Thus, we introduce a regularization loss that is similar the losses at the initialization stage to prevent the network from generating unrealistic results or overfitting. The regularization loss contains a texture term:
\begin{equation}
   L_{RT} = \sum_{j=1}^b\| \sigma_{j}\odot (T - T_{j})\|_1,
   \label{eqn:lrt}
\end{equation}
a coordinate term:
\begin{equation}
   \begin{aligned}
      L_{RC} =& \sum_{k=1}^{24}(\|S^k\odot (C^{2k}-C^{*2k})\|_1 + \\ & \|S^k\odot (C^{2k+1}-C^{*2k+1})\|_1),
   \end{aligned}
   \label{eqn:lrc}
\end{equation} 
and a body part mask term that ensures $S$ do not deviates too much from $S^*$ output by the DensePose model on the foreground mask:
\begin{equation}
   L_{RM} = L_{CE}((1-S^{*25}) \odot S, (1-S^{*25}) \odot S^*),
\end{equation}

The total loss can be expressed as follows:
\begin{equation}
   L = \lambda_I L_I + \lambda_M L_M + \lambda_{RT} L_{RT} + \lambda_{RC} L_{RC} + \lambda_{RM} L_{RM},
\end{equation}
where $\lambda$'s are the weights balancing the contribution of each loss term. Note that we do not add adversarial loss during image reconstruction as we find it would add instability to the training of geometry generator.

\subsection{Few-shot Fine-tuning}
The model trained at our multi-video training stage can be readily used for generating human motion transfer results on unseen people and poses. Fine-tuning on a few source images at test time is an optional step that greatly improves the quality of the synthesized images.

Since we only need to generate one texture map $T$ for one person at test time, fine-tuning of texture map seldom overfits. If the texture map gets closer to the texture of source images, the texture of synthesized human at other pose becomes more photo-realistic. 

We fine-tune the embedding $t$ in \Cref{eqn:embed} to produce a smooth and realistic texture map. Compared with directly optimizing texture map $T$ as in \cite{texturedavatars}, fine-tuning embedding causes few artifacts to the texture map and it is able to hallucinate incomplete textures as shown in \Cref{fig:texture_compare}. During test time fine-tuning, we first initialize $t$ by averaging the embedding of source textures ${T}_{1}, ... , {T}_{n}$. And generate complete image $\widehat I$ with \Cref{eqn:fgbg} and fine-tune $t$ by minimizing:
\begin{align}
    &\widehat L_{I} = \|\widehat{I} - I^*\|_1 + \sum_{v=1}^N \|\Phi^v (\widehat{I}) - \Phi^v (I^*)\|_1 \\ 
    &L_{test} = \lambda_I \widehat L_{{I}} + \lambda_M L_M + \lambda_{RC} L_{RC} + \lambda_{RM} L_{RM}.
\end{align}
We do not add $L_{RM}$ as we do not hope to constrain the optimization of texture map at test time. Meanwhile, we also fine-tune the geometry generator $G_\phi$ and through $L_{test}$ for few steps since the geometry generator can already generalize pretty well to unseen person geometries and fine-tuning for a long period may lead to overfitting. We also fine-tune background $B$ through $L_{test}$ as it can eliminate artifacts for merging backgrounds.

\section{Experiments}

\begin{figure*}[t!]
    \includegraphics[width=\linewidth]{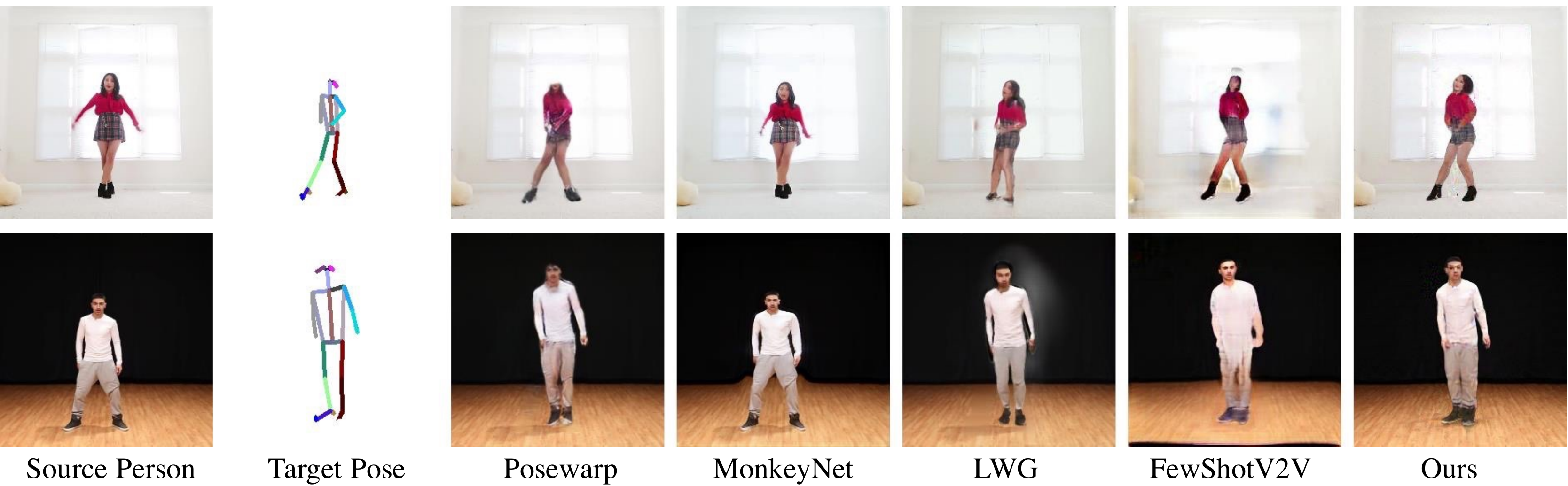}
   \caption{Qualitative comparison of our method to state-of-the-art methods. \textit{More video results can be found in the supplementary material.}}
   \label{fig:compare}
   \vspace{-0.3cm}
\end{figure*}

\begin{figure*}[t]
 \centering
    \includegraphics[width=\linewidth]{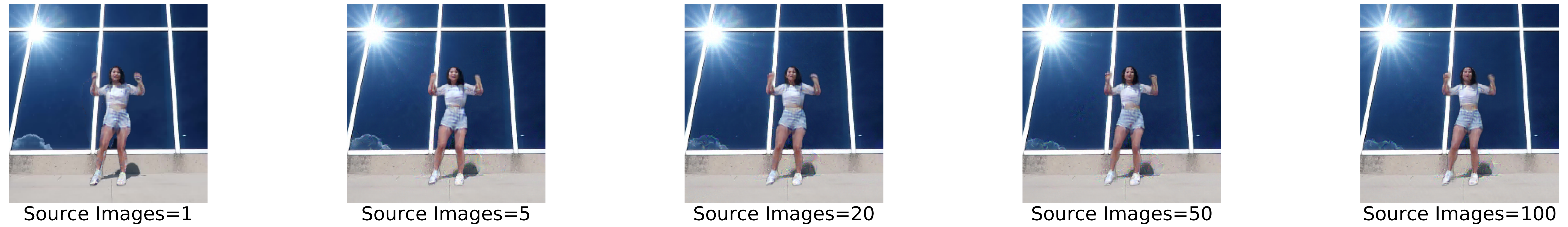}
  \caption{Synthesized human with different number of source images. }
  \label{fig:change_number}
  \vspace{-0.4cm}
\end{figure*}

\begin{figure*}[t!]
 \centering
    \includegraphics[width=1.0\linewidth]{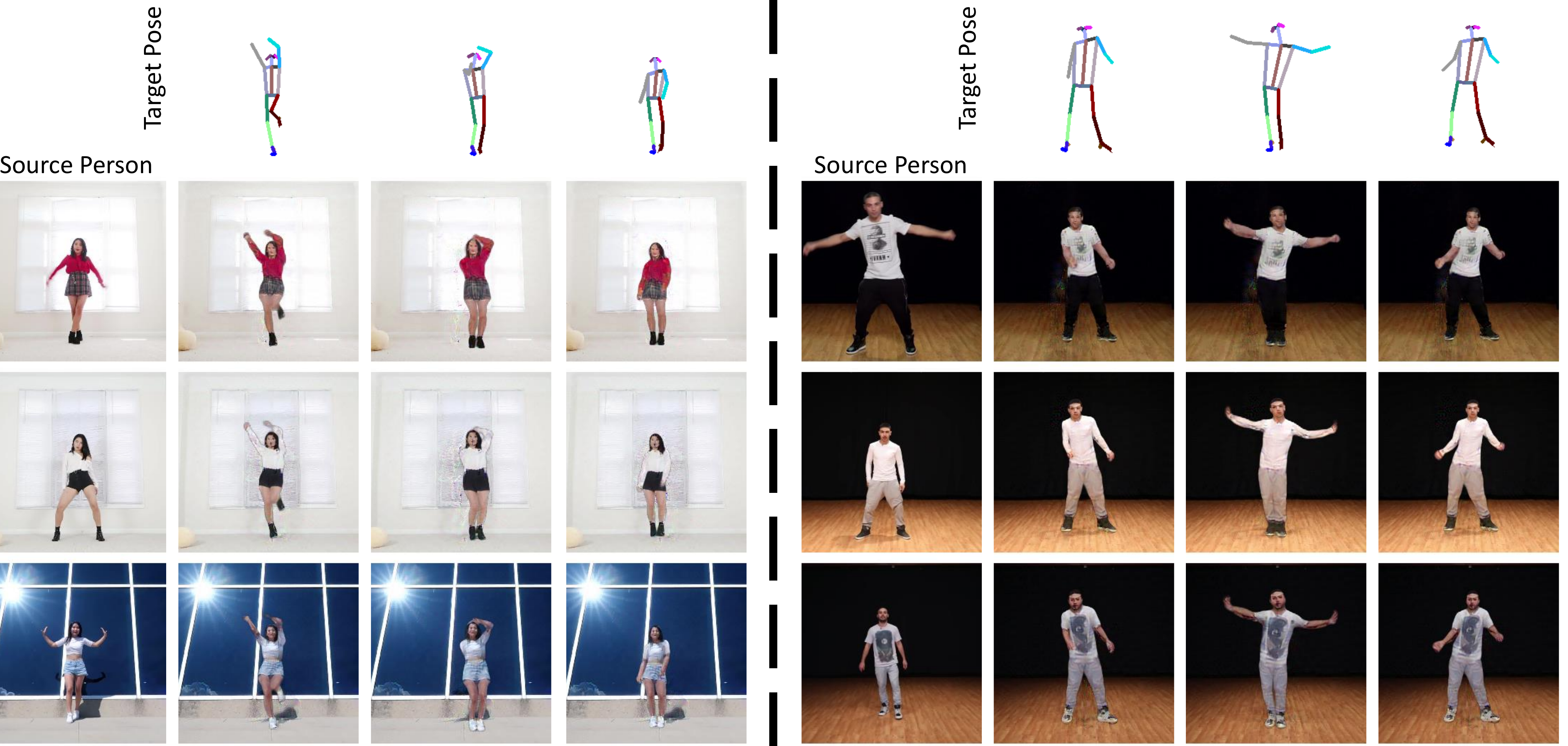}
   \caption{More examples of our method. Our method can generate personalized geometry and realistic texture for a wide variety of source persons. \textit{More video results can be found at \url{https://youtu.be/ZJ15X-sdKSU}.}}
   \label{fig:example}
  \vspace{-0.5cm}
\end{figure*}

\noindent\textbf{Dataset.} We collect 62 solo dance videos with almost static background from YouTube. The videos contain several subjects with different genders, body shapes, and clothes (examples can be found in Figure \Cref{fig:example}). Each video is trimmed into a clip lasting about 3 minutes. We further divide them into training and test set with no overlapping subjects. Training set contains 50 videos with 283,334 frames and test set contains 12 videos with 70,240 frames. 

\noindent\textbf{Preprocess.} For each frame $I$ in the dataset, we crop the center part of the image and extract the 25 body joints with OpenPose \cite{openpose, openpose2}. The joints are connected to form a ``stickman'' image as the input pose $P$. UV coordinate $C^*$ and mask $S^*$ are extracted with DensePose model \cite{densepose}. DensePose just gives a coarse segmentation of the human. We use Deeplab V3 \cite{humanseg} to get foreground mask $m$ separate foreground human image $\widetilde{I}$. Source texture $\widetilde{T}$ is produced by warping the image $I$ according to $C^*$.

\noindent\textbf{Implementation Details.} All video frames are resized to 256$\times$256 while the size of body part texture map is set to $128\times128$. The input pose is a ``stickman" image with 26 channels and each channel contains one ``stick" of the pose. The encoder and decoder for geometry and texture generators are built on basic convolution-relu-norm blocks, and we include more detailed illustrations in the supplementary material. Geometry generator contains about 60M parameters and that of texture generator is around 34M.

Both generators are training using Adam \cite{kingma2014adam} optimizer with $(\beta_1, \beta_2) = (0.5, 0.999)$. Learning rate starts at $0.0002$. The initialization stage lasts for 10 epochs and the learning rate decays half at the 5th epoch. We train the multi-video stage for 15 epochs with learning decaying half at the 5th and the 10th epoch. At test time, we randomly select 20 images from one video as the source images. The number of fine-tuning steps is 40 for geometry generator and is 300 for the texture embedding. We generate the background $B$ by directly merging visible background from source images and fill the left invisible parts with deepfillv2 \cite{deepfillv2}.

\noindent\textbf{Compared Methods.} We compare our method with state-of-the-art human motion transfer approaches:
Posewarp \cite{posewarp}, MonkeyNet \cite{monkeynet}, FewShotV2V \cite{fewshotvid} and Liquid Warping GAN (LWG) \cite{liu2019liquid}. We use the source code provides by the authors to train the model. For those providing pre-trained models (Posewarp, LWG), we use it as initialization and train on our dataset for fair comparison. Otherwise, we train the model from scratch. At test time, we fine-tune all these models on the source images. As only FewShotV2V accepts multiple inputs, for other methods, the source image is chosen to be the first one.

\begin{table}[t]
      \setlength\tabcolsep{2.7pt}
   \caption{Quantitative evaluation metrics on our test set. $\uparrow$ represents higher is better, and $\downarrow$ means lower is better.}
      \begin{tabular}{l c c c c}
         \hline
         \multirow{2}*{Methods} & 
         \multicolumn{2}{c}{Reconstruction} & \multicolumn{2}{c}{Motion Transfer} \\
         \cmidrule(r){2-3} \cmidrule(r){4-5}
         & SSIM$\uparrow$ & LPIPS$\downarrow$ & FReID$\downarrow$ & PoseError$\downarrow$\\          \hline
         Posewarp~\cite{posewarp} & 0.808 & 0.175 & 5.73 & 12.46 \\
         MonkeyNet~\cite{monkeynet} & 0.763 & 0.234 & 13.40 & 41.49 \\
         LWG~\cite{liu2019liquid} & 0.760 & 0.238 & 9.99 & 14.45 \\
         FewShotV2V~\cite{fewshotvid} & 0.694 & 0.332 & 8.24 & 10.50\\
         \hline 
         Ours$_\text{w/o Fine-tuning}$ & 0.837 & 0.166 & 4.91 & \textbf{6.48} \\
         Ours$_\text{w/ Fine-tuning}$ & 0.861 & 0.157 & 3.78 & 6.58\\ 
         \hline
         Ours$_\text{Direct Merge}$
         & 0.843 & 0.183 & 4.40 & 8.09 \\
         Ours$_\text{Texture Map}$ & \textbf{0.881} & \textbf{0.151} & \textbf{3.23} & 7.27 \\
         \hline
      \end{tabular}

   \label{tab:quantity}
    \vspace{-0.2cm}
\end{table}

\begin{table}[t]
   \centering
 \setlength\tabcolsep{8pt}
    \caption{User study of human motion transfer. The numbers indicate the percentage of clips that the users prefer our method to each of the competing method. Chance is 50\%.}
      \begin{tabular}{c c c c}
         \hline
         Posewarp & MonkeyNet & LWG & FewShotV2V \\\hline
         98.40\% & 99.47\% & 87.50\% & 90.43\% \\
         \hline
      \end{tabular}

  \label{tab:user_study}
    \vspace{-0.4cm}
\end{table}

\subsection{Quantitative Comparisons}

\noindent\textbf{Evaluation Metrics.} For each video in the test set, we consider two evaluation settings: reconstruction and motion transfer. For the reconstruction, we set the pose sequence from the video of the source person as the target motion. The generator is asked to reconstruct images to be the same as the ground-truth video. We compare the similarity of the synthesized and ground-truth image using SSIM \cite{ssim} and Learned Perceptual Similarity (LPIPS) \cite{lpips}. For motion transfer, we extract the pose sequence from one video in the test set and let a person in other videos imitate the pose sequence. We compare the quality of the images in two aspects: Pose Error \cite{everybodydancenow,fewshotvid} and FReID \cite{liu2019liquid}. Pose Error is the average L2 distance (in pixels) of 2D pose keypoints between of synthesized human and target pose, which estimates the accuracy of the transferred motion. FReID is a Fréchet Distance \cite{fid} on a pre-trained person-reid model \cite{reid}, which measures the quality of generated appearance. The results are shown in \Cref{tab:quantity}. Our method achieves higher similarity in reconstruction and lower discrepancy of pose and appearance in motion transfer.

\noindent\textbf{User Study} We run all the methods on 30 randomly chosen 5-second video clips for human motion transfer and ask 8 people to perform user study. In each trial, given the motion transfer results of the five methods on the same video clip, the users are asked to select the method with the best generation quality. The percentage of trials that our method is preferred is shown in \Cref{tab:user_study}. We can find that our method is favored in most of the clips.

\subsection{Qualitative Comparisons}

\Cref{fig:compare} visually compares our method with others. Our method outperforms state-of-the-art methods in terms of the image quality and pose accuracy. Posewarp fails to construct some parts of the body. FewShotV2V cannot generalize to new person with small number of training videos (their paper used 1500 videos). It only outlines the appearance of source person with plenty of artifacts. While LWG can synthesize a person with regular shapes, it is not able to generate person with complicated shape such as dress and long hair, as detailed shape information cannot be modeled by HMR \cite{hmr} used in LWG. Furthermore, HMR cannot accurately estimate the target pose, making generated results temporally discontinuous. Our method is capable of modeling the complicated personalized geometry of each person and preserve detailed appearance.

\Cref{fig:example} shows more examples of motion transfer generated by our method, which produces geometry with accurate pose and personalized shape details for a large variety of people. Besides, the texture is well preserved for the synthesized human, resulting in high-fidelity motion transfer. 

\begin{table}[t]
  \setlength\tabcolsep{5.4pt}
   \centering
       \caption{Performance with varied numbers of source images.}
      \begin{tabular}{l c c c c c}
         \hline
         Source Images & 1 & 5 & 20 & 50 & 100 \\\hline
         SSIM$\uparrow$ & 0.845 & 0.852 & 0.861 & 0.867 & 0.870 \\
         LPIPS$\downarrow$ & 0.167 & 0.174 & 0.157 & 0.158 & 0.153\\ \hline
         FReID$\downarrow$ & 4.29 & 3.97 & 3.78 & 3.60 & 3.41 \\
         Pose Error$\downarrow$ & 6.86 & 6.68 & 6.58 & 6.53 & 6.58\\
         \hline
      \end{tabular}

  \label{tab:change_number}
   \vspace{-0.4cm}

\end{table}


\subsection{Ablation Study}

\noindent\textbf{Number of Source Images.} We vary the number of source images at test time from 1 to 100 and present the results in \Cref{fig:change_number} and \Cref{tab:change_number}. Our method achieves high quality motion transfer with only one source image, and the generation quality improves as more source images are utilized.

\begin{figure}[t]
    \centering
    \subfigure[Direct Merge]{
        \begin{minipage}[t]{0.22\linewidth}
            \centering
            \includegraphics[width=\linewidth]{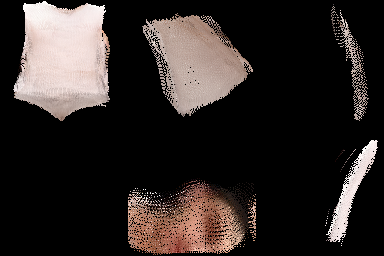}
            \includegraphics[width=\linewidth]{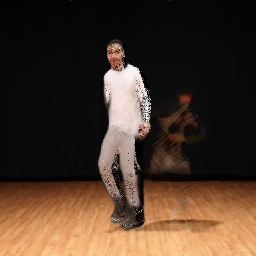}
        \end{minipage}
    }
    \subfigure[No Fine-tune]{
        \begin{minipage}[t]{0.22\linewidth}
            \centering
            \includegraphics[width=\linewidth]{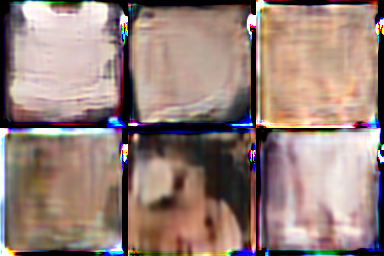}    
            \includegraphics[width=\linewidth]{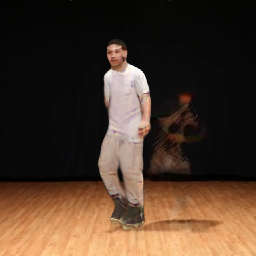}\label{fig:no-finetune}

        \end{minipage}
    }
    \subfigure[Texture Map]{
        \begin{minipage}[t]{0.22\linewidth}
            \centering
            \includegraphics[width=\linewidth]{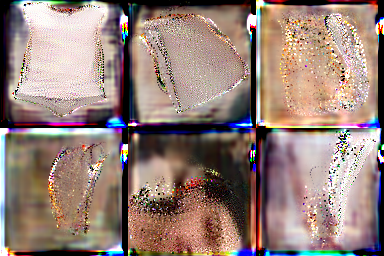}   
            \includegraphics[width=\linewidth]{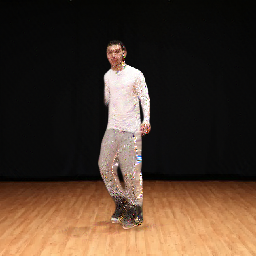} 
        \end{minipage}
    }
    \subfigure[Embedding]{
            
        \begin{minipage}[t]{0.22\linewidth}
            \centering
            \includegraphics[width=\linewidth]{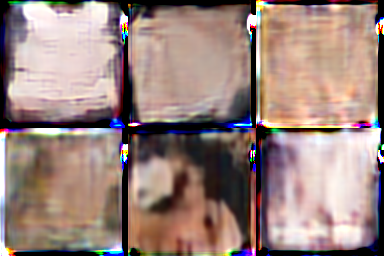} 
            \includegraphics[width=\linewidth]{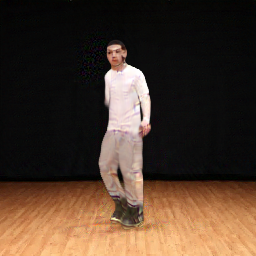}\label{fig:embedding} 
        \end{minipage}
    }
    \caption{Texture maps of different fine-tuning schemes.}
    \label{fig:texture_compare}
    \vspace{-0.6cm}
\end{figure}

\noindent\textbf{Texture Map.} \Cref{fig:texture_compare} shows the learned texture maps and human images of four strategies: directly averaging the source textures, without fine-tuning, fine-tuning texture map, and  fine-tuning the embedding. Quantitative metrics are shown in \Cref{tab:quantity} (Ours$_\text{w/o Fine-tuning}$ means \cref{fig:no-finetune}, Ours$_\text{w/ Fine-tuning}$ means \cref{fig:embedding}). Although fine-tuning texture map has better quantitative metrics, the noises of the texture make the synthesized video quite unnatural for human eyes.
Texture generator fills in the invisible parts of the source texture and fine-tuning the embedding further improves the quality without suffering from visual artifacts.

\vspace{-0.2cm}
\section{Conclusion}

We proposed a novel method for few-shot human motion transfer, which decouples the task into generation of personalized geometry and texture. We designed a geometry generator that can extract shape information from  source person images  and inject it into generating personalized geometry of the source person in the target pose. In addition, a texture generator merges source textures and fills in invisible texture map. 
Extensive experiments demonstrate that the proposed method outperforms previous approaches for synthesizing realistic human motion. We see our  method may be limited in coping with non-rigid moving parts like whipping hair or shaking dress. One future direction is to have the geometry generator take multiple continuous frames as input, and learn the temporally consistent motion.

{\small
\bibliographystyle{ieee_fullname}
\bibliography{egbib}
}

\end{document}


\title{Supplementary Material for Few-Shot Human Motion Transfer by \\ Personalized Geometry and Texture Modeling}


\maketitle

In this supplementary material, we first describe the detailed network architecture of our geometry generator and texture generator. Meanwhile, the choices of hyper-parameters are provided. We present extra experimental results, including a user study and more visual comparisons. We also perform an extra experiment on large number of source images to demonstrate that our method can synthesize realistic human as a single person model. A supplementary video that contains visual results of our method and the compared baselines is available at \url{https://youtu.be/ZJ15X-sdKSU}.

\section{Details of Experiments}

\subsection{Network Architecture}

\noindent\textbf{Notations for the architecture}
\begin{itemize}
    \item Conv($a, b, c, d$): 2d-convolution operation with input channel $a$, output channel $b$, kernel size $c\times c$ and stride $d$;
    \item CRN($a, b, c, d$): Conv($a, b, c, d$)-relu-normalization sequence;
    \item ResBlk($a$): residual block used in ResNet with input and output channel $a$;
    \item Upsample($a$): bilinear interpolation to upsample the input feature to $a$ times of its original spatial resolution.
\end{itemize}
The padding strategy for convolution is ``SAME". We use instance normalization since our batch size is limited by the memory of GPU and batch normalization does not work well when the batch size is small. The $O_i$'s and $I_i$'s denote the position where we merge the feature of source images and target pose as shown in Figure 2(a) in our main paper. The output at $O_i$ will be merged with attention and sent into the corresponding $I_i$.

\noindent\textbf{Image Context Encoder $E_I$:} CRN(3, 32, 7, 1) $\xrightarrow{}$ CRN(32, 64, 3, 2) $\xrightarrow{O_1}$ CRN(64, 128, 3, 2) $\xrightarrow{O_2}$ CRN(128, 128, 3, 2) $\xrightarrow{O_3}$ CRN(128, 128, 3, 2) $\xrightarrow{O_4}$

\noindent\textbf{Pose Attention Encoder $E_W$:} CRN(26, 32, 7, 1) $\xrightarrow{}$ CRN(32, 64, 3, 2) $\xrightarrow{O_1}$ CRN(64, 128, 3, 2) $\xrightarrow{O_2}$ CRN(128, 128, 3, 2) $\xrightarrow{O_3}$ CRN(128, 128, 3, 2) $\xrightarrow{O_4}$

\noindent\textbf{Target Pose Encoder $E_P$:} CRN(26, 32, 7, 1) $\xrightarrow{}$ CRN(32, 64, 3, 2) $\xrightarrow{O_1}$ CRN(64, 128, 3, 2) $\xrightarrow{O_2}$ CRN(128, 128, 3, 2) $\xrightarrow{O_3}$ CRN(128, 128, 3, 2) $\xrightarrow{O_4}$ 10$\times$Resblk(128)

\noindent\textbf{Mask Branch of Decoder $D_G$:} $\xrightarrow{I_4}$ Upsample(2) $\xrightarrow{}$ CRN(256, 128, 5, 1) $\xrightarrow{I_3}$ Upsample(2) $\xrightarrow{}$ CRN(256, 128, 5, 1) $\xrightarrow{I_2}$ Upsample(2) $\xrightarrow{}$ CRN(256, 64, 5, 1) $\xrightarrow{I_1}$ Upsample(2) $\xrightarrow{}$ CRN(128, 32, 5, 1) $\xrightarrow{}$ Conv(32, 25, 7, 1)

\noindent\textbf{Coordinate Branch of Decoder $D_G$:} $\xrightarrow{I_4}$ Upsample(2) $\xrightarrow{}$ CRN(256, 128, 5, 1) $\xrightarrow{I_3}$ Upsample(2) $\xrightarrow{}$ CRN(256, 128, 5, 1) $\xrightarrow{I_2}$ Upsample(2) $\xrightarrow{}$ CRN(256, 64, 5, 1) $\xrightarrow{I_1}$ Upsample(2) $\xrightarrow{}$ CRN(128, 32, 5, 1) $\xrightarrow{}$ Conv(32, 48, 7, 1)

\noindent\textbf{Texture Encoder $E_T$:} CRN(3, 64, 7, 1) $\xrightarrow{}$ CRN(64, 128, 3, 2) $\xrightarrow{}$ CRN(128, 256, 3, 2) $\xrightarrow{}$ CRN(256, 512, 3, 2)

\noindent\textbf{Texture Encoder $D_T$:} 6$\times$Resblk(512) $\xrightarrow{}$ Upsample(2) $\xrightarrow{}$ CRN(512, 256, 3, 1) $\xrightarrow{}$ Upsample(2) $\xrightarrow{}$ CRN(256, 128, 3, 1) $\xrightarrow{}$ Upsample(2) $\xrightarrow{}$ CRN(128, 64, 3, 1)  $\xrightarrow{}$ Conv(64, 3, 7, 1)

\subsection{Hyperparamters}

For $\lambda$'s mentioned in the Equation (19), we empirically set $\lambda_I=1, \lambda_M =1, \lambda_{RT}=1, \lambda_{RC}=20, \lambda_{RM} = 0.8$. The $\lambda$'s mentioned in the Equation (21) are the same.

We use Adam optimizer with $(\beta_1, \beta_2) = (0.5, 0.999)$ at all initialization, multi-video training and fine-tuning stages. For initialization and multi-video training stages, learning rate starts at $0.0002$. The initialization stage lasts for 10 epochs and the learning rate decays half at the 5th epoch. We train the multi-video stage for 15 epochs with learning decaying half at the 5th and the 10th epoch. At the test time, we randomly select 20 images from one video as the source images. The number of fine-tuning steps is 40 for geometry generator and 300 for the texture embedding and the background. We set the learning rate to be $0.0002$ for the geometry generator and $0.005$ for the texture embedding and the background. We train the generators on two GPUs with their memories fully utilized. We set batch size to be 16 for initialization of geometry generator, 8 for initialization of texture generator and 10 for the multi-video training stage. We fine-tune the model on one GPU and the batch size is 6.




\section{More Experiments}
We include more visual results in the supplementary video. 



\begin{figure*}[t!]
    \centering
    \includegraphics[width=\linewidth]{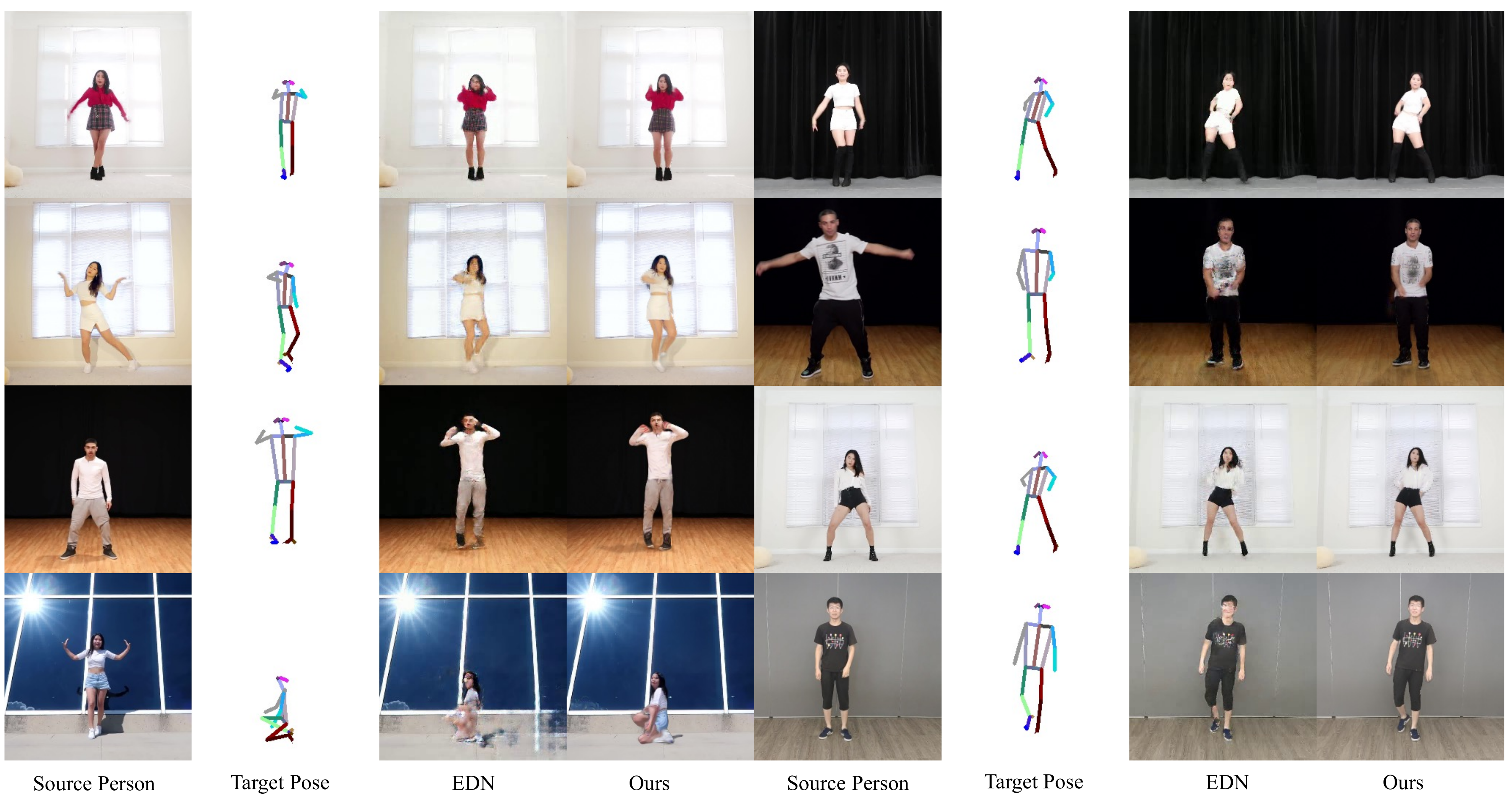}
    \caption{Visual comparison between our method and EDN \cite{everybodydancenow} with sufficient number of source images. Our method generates more realistic human. }
    \label{fig:compare_EDN}
\end{figure*}

\begin{figure*}[t!]
    \centering
    \subfigure[20 Source Images]{
        \begin{minipage}[t]{0.48\linewidth}
            \centering
            \includegraphics[width=0.48\linewidth]{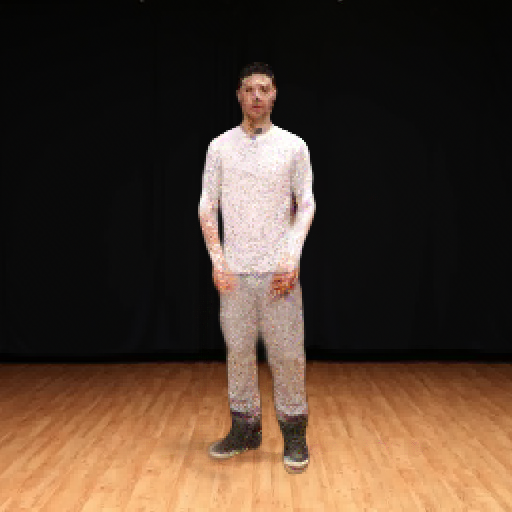}
            \includegraphics[width=0.48\linewidth]{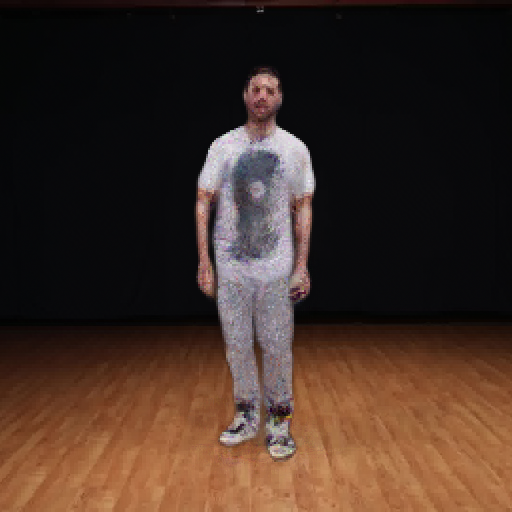}
            \includegraphics[width=0.48\linewidth]{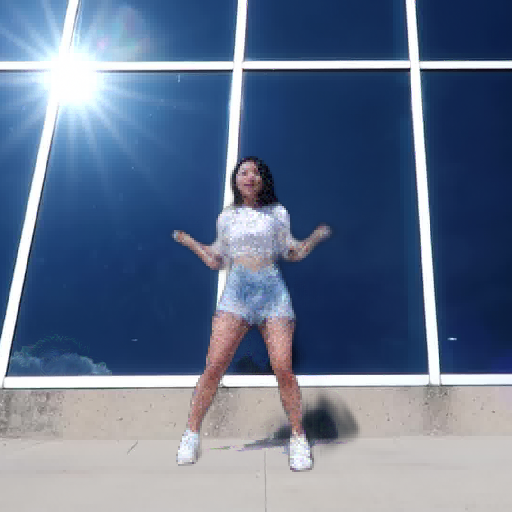}
            \includegraphics[width=0.48\linewidth]{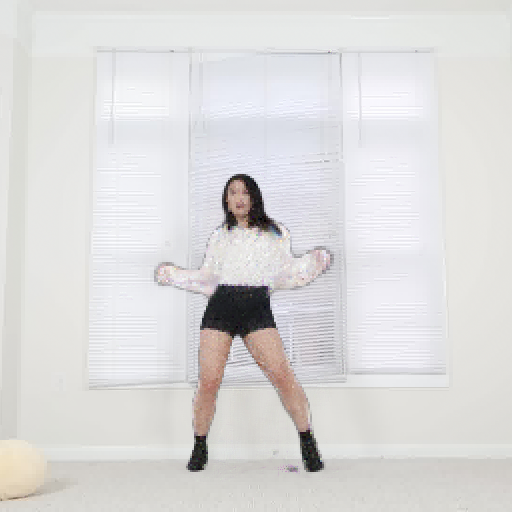}
            \label{fig:HD_images_1}
        \end{minipage}
    }    
    \subfigure[Sufficient Source Images]{
        \begin{minipage}[t]{0.48\linewidth}
            \centering
            \includegraphics[width=0.48\linewidth]{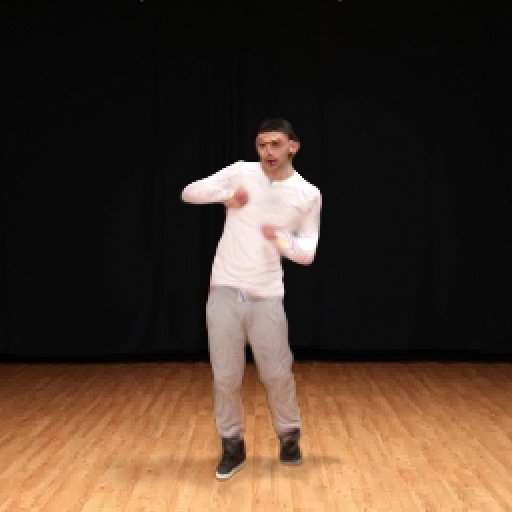}
            \includegraphics[width=0.48\linewidth]{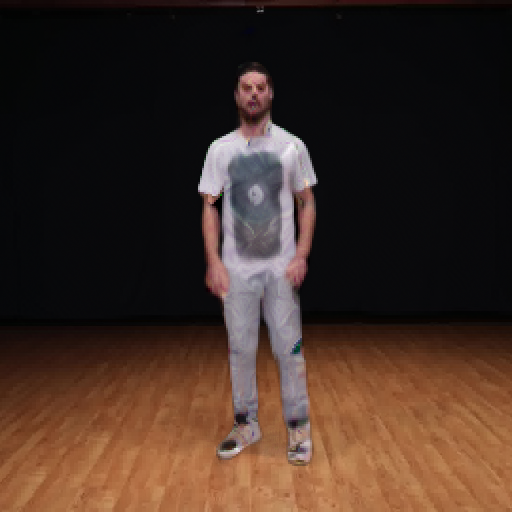}
            \includegraphics[width=0.48\linewidth]{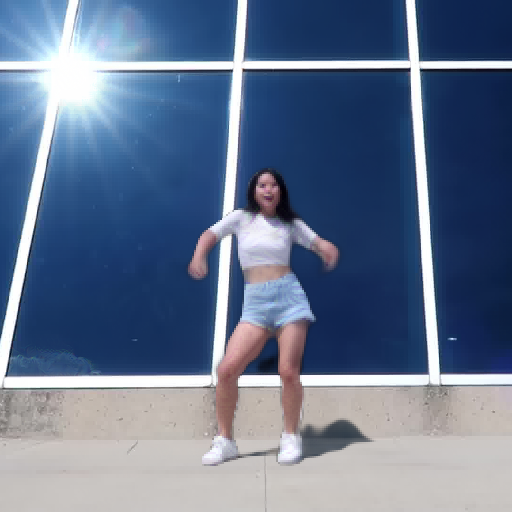}
            \includegraphics[width=0.48\linewidth]{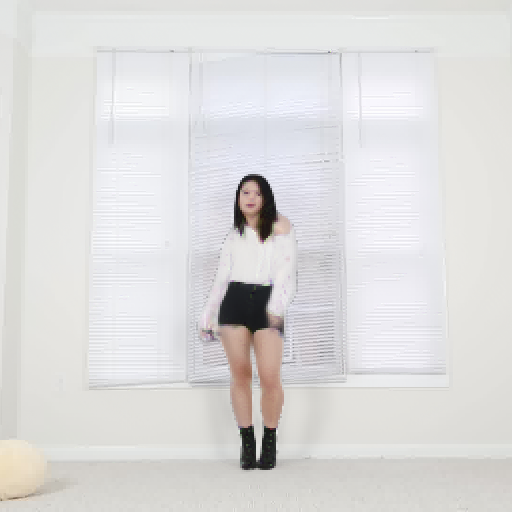}
            \label{fig:HD_images_2}
        \end{minipage}
    }    
    \caption{Examples of $ 512\times 512$ HD images by up-sampling the UV maps and fine-tune the texture on high resolution source images.}
    \label{fig:HD_Images}
\end{figure*}

\subsection{Sufficient Number of Source Images}

As the number of source images grows, our method is able to generate realistic human motion as single person model. We fine-tune our model on all available images of the source person. The number of frames for each source person is about 4,000 and slightly varies across different source persons. We obtain the best \textbf{FReID 2.33} and \textbf{pose error 6.34}. The visual results are shown in \Cref{fig:single_1} and \Cref{fig:single_2}. We observe that high quality images are effectively generated with fine texture details and accurate target pose. For example, realistic textures on the T-shirt and shoes are well preserved in the third example of \Cref{fig:single_2}. Interestingly, in the second example of \Cref{fig:single_2}, shadows can be accurately generated as the fine-tuning process enables the geometry generator to model the shadow as one part of the human.

\noindent\textbf{Comparison with EDN \cite{everybodydancenow}.} We compare our method with EQN when the number of source images is sufficient. We remove the face generator from EDN and train individual model for each person using all images of the source person. In the task of motion transfer, we achieve \textbf{FReID 5.21} and \textbf{pose error 9.45}, which is inferior to our method. \Cref{fig:compare_EDN} shows more visual comparisons between our method and EDN. Our proposed method achieves higher synthesis quality than EDN.

\subsection{Generating HD Images}

Our method is able to generate HD images using generators trained on low resolution images.  The resolution of the neural rendering mainly relies on the resolution of the texture map. As the high resolution UV map is almost continuous, we can simply up-sample the UV map with bilinear interpolation. And the high resolution texture can be obtained by directly fine-tuning the texture map on HD images.  \Cref{fig:HD_Images} shows some examples of the generating HD images on $512 \times 512$ resolution with few or sufficient number of source images.

\subsection{More Qualitative Comparison}

\begin{figure*}[h!]
    \centering
    \includegraphics[width=\linewidth]{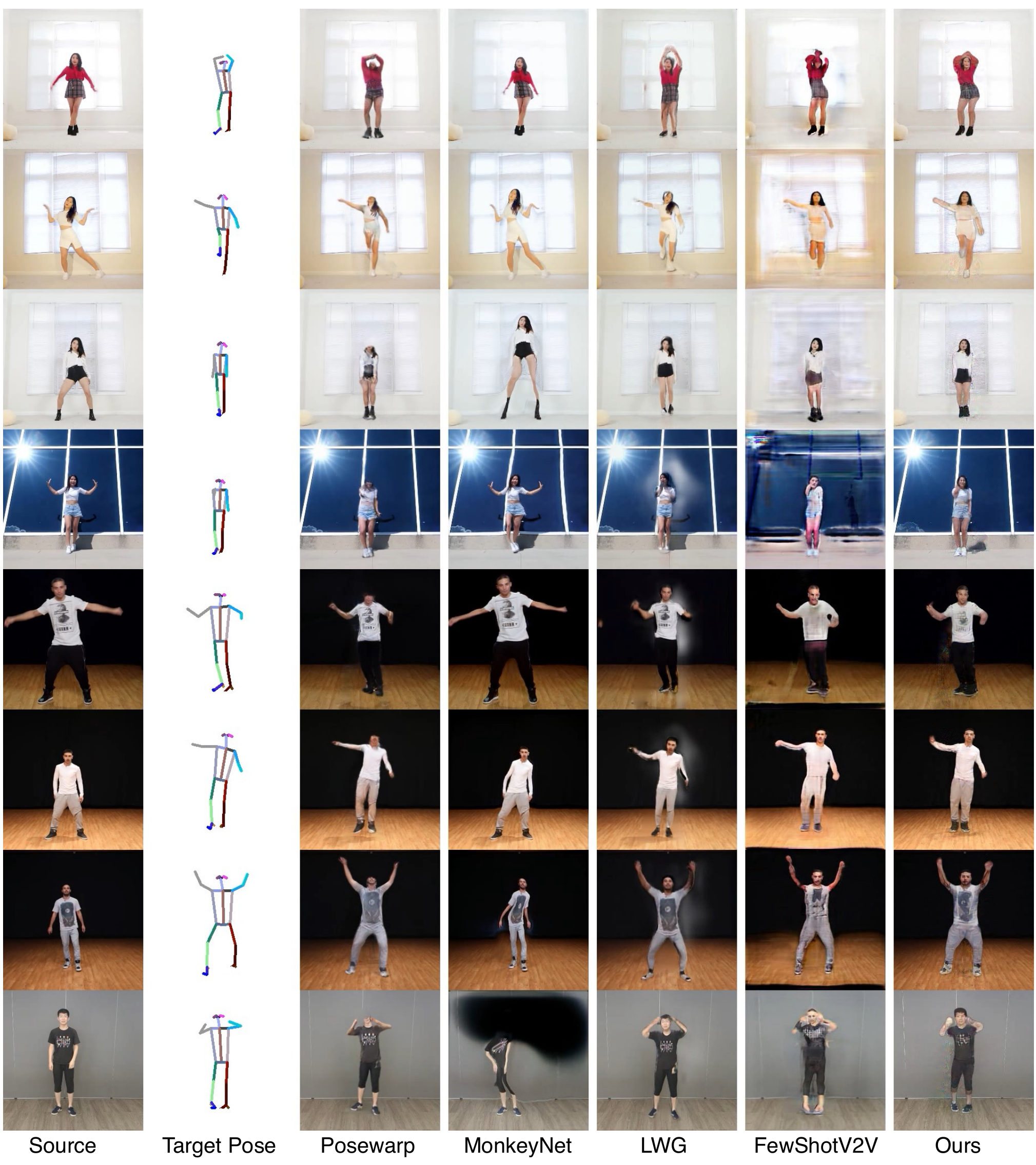}

    \caption{More visual comparison between our method and other competing methods. Supplementary video contains the corresponding video results.}
    \label{fig:compare}
\end{figure*}

\begin{figure*}[h!]
    \centering
    \includegraphics[width=\linewidth]{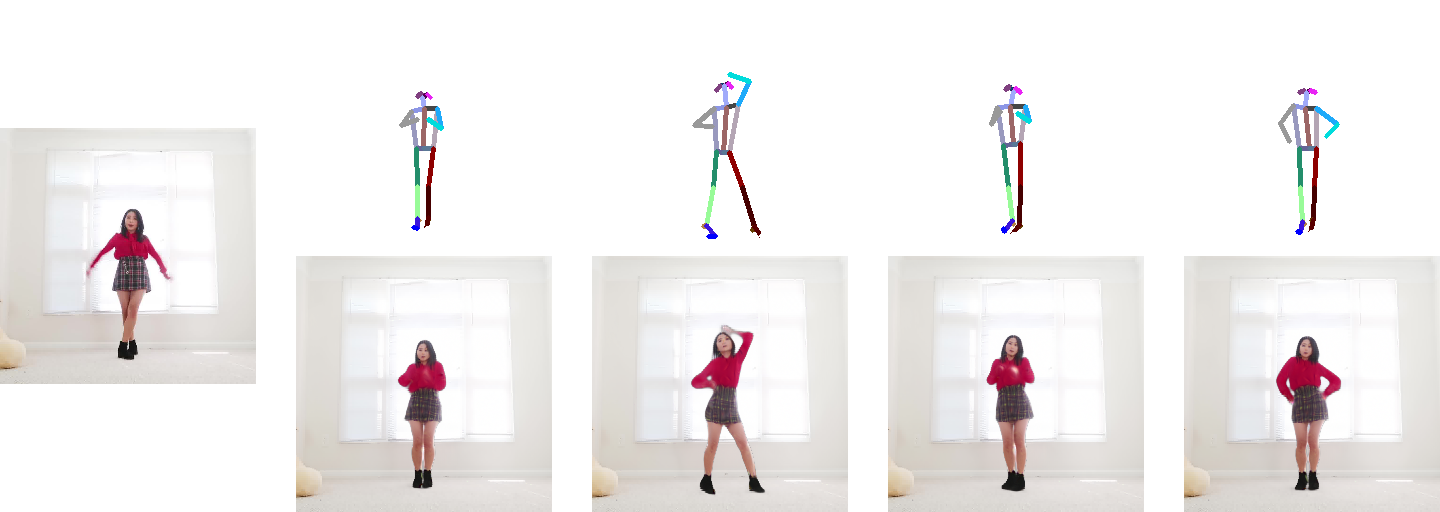}
    \includegraphics[width=\linewidth]{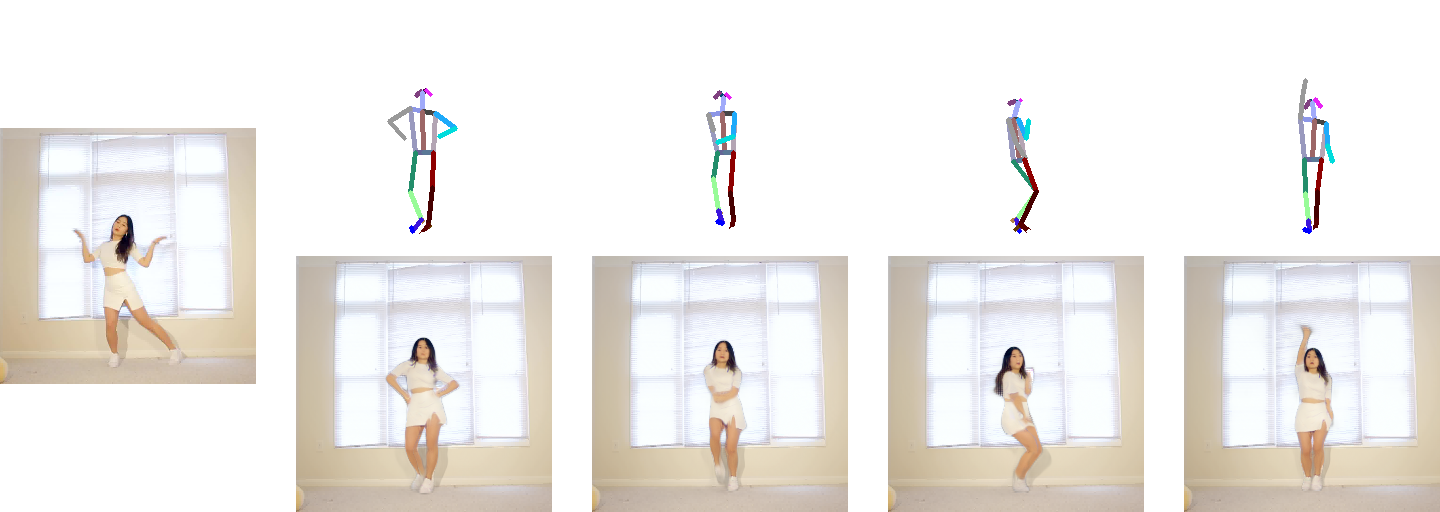}
    \includegraphics[width=\linewidth]{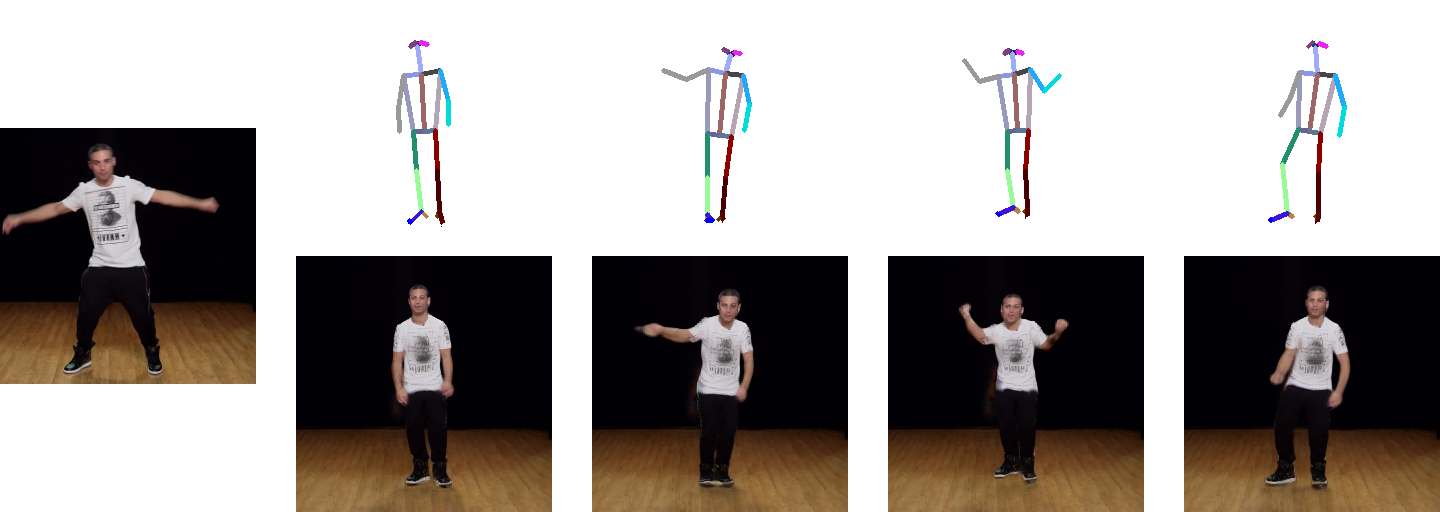}

    \caption{The synthesized motion of our method with sufficient number of source images. The first column represents source person and the other columns are the synthesized human in the corresponding pose. Video results are included in the supplementary video.}
    \label{fig:single_1}
\end{figure*}

\begin{figure*}[h!]
    \centering
    \includegraphics[width=\linewidth]{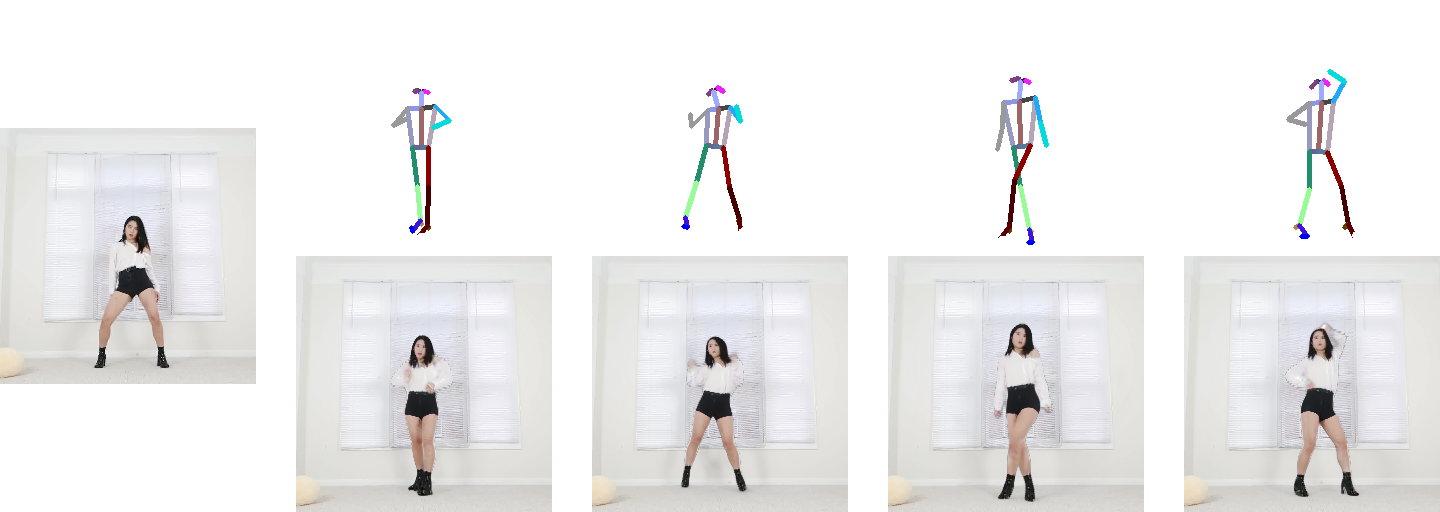}
    \includegraphics[width=\linewidth]{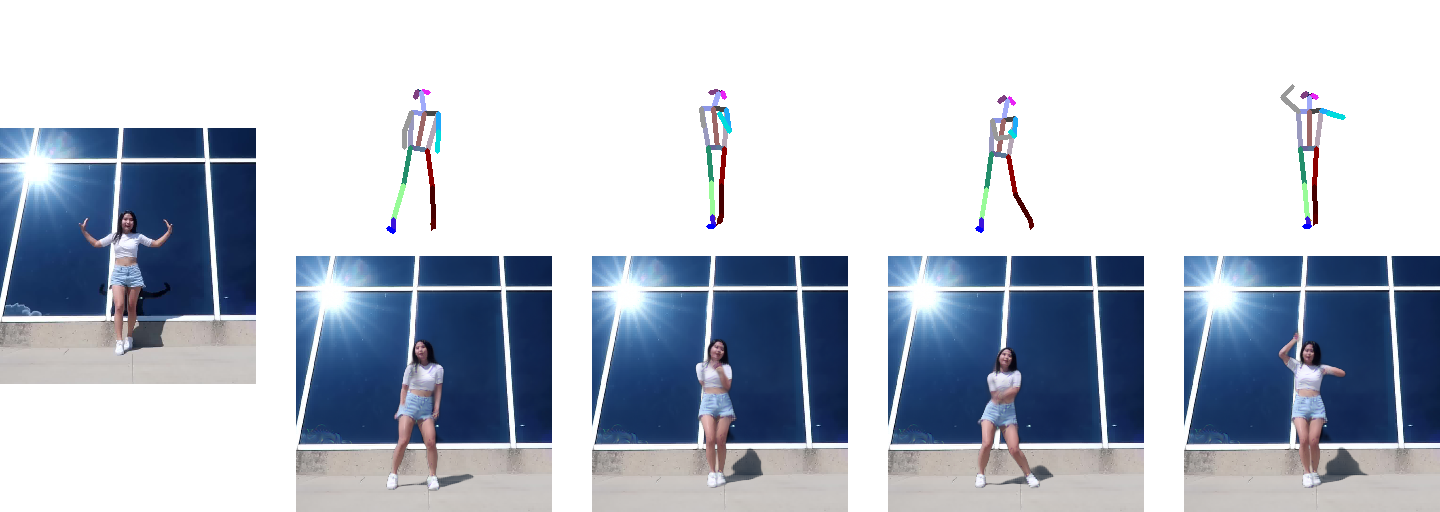}
    \includegraphics[width=\linewidth]{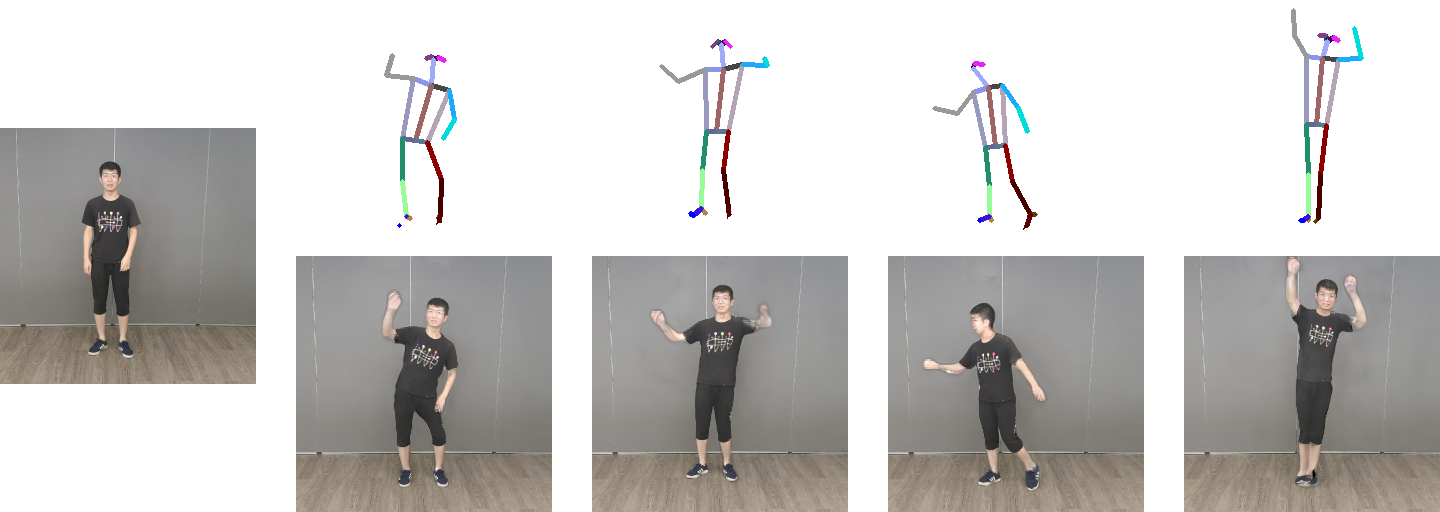}

    \caption{The synthesized motion of our method with sufficient number of source images. The first column represents source person and the other columns are the synthesized human in the corresponding pose. Video results are included in the supplementary video.}
    \label{fig:single_2}
\end{figure*}

\Cref{fig:compare} includes more synthesized images of human motion transfer of different methods in the few-shot setting. We give more detailed analysis of each competing method as follows.

\noindent\textbf{Posewarp \cite{posewarp}.} Posewarp segments each body part by the predefined masks and uses affine transformation to transfer body parts from the source pose to the target pose. However, affine transformation only produces coarse body parts in the target pose. The generative network at later stage has limited power and cannot recover all the details of the source person. Besides, the predefined mask cannot generalize to different persons with different shapes. Therefore, Posewarp may lose some parts of the body (\eg head regions in \Cref{fig:compare}). The resulting artifacts are consistent with the what \cite{fewshotvid,liu2019liquid} report in their experiments.

\noindent\textbf{MonkeyNet \cite{monkeynet}.} MonkeyNet does not use 2D pose keypoints extracted by off-the-shelf models. Instead, it uses the keypoints detected by itself and moves these keypoints to generate motion for the source person. However, these keypoints are necessarily aligned with the pose and moving these keypoints does not change the pose of the source person. It only adds strange distortion to the source image while makes the pose stay the same. Similar observations are also made in the experimental comparison in \cite{fewshotvid}.

\noindent\textbf{LWG \cite{liu2019liquid}.} While LWG can produce realistic images for persons with regular shapes (the 5th-8th rows of \Cref{fig:compare}), it is not able to generate persons with complicated shapes such as dress and long hair (the 1st-4th rows of \Cref{fig:compare}). This is because LWG is built on top of SMPL models predicted by HMR \cite{hmr}, and SMPL is a skinned 3D human model that does not model human's clothes and hair. Furthermore, HMR might be inaccurate when estimating the target pose especially when the background is cluttered or there is motion blur, making the generated results temporally discontinuous (see supplementary video for temporal comparison).

\noindent\textbf{FewShotV2V \cite{fewshotvid}.} FewShotV2V only outlines the human in the target pose with plenty of artifacts. Although FewShotV2V uses SPADE module to generate different weights for different person, the module requires a lot of training subjects (1,500 videos as described in the paper) so that it can learn how to generate weights for different persons. And the 50 videos we use are not enough for FewShotV2V to learn good generalization. Our method, in contrast, decouples the generation of geometry and texture and it can generalize to a new person with only a small number of training videos.

Compared with these methods, our method obtains the most visual appealing results without presenting the aforementioned artifacts.

\subsection{Texture Map}
\begin{figure*}[t]
    \centering
    \subfigure[Direct Merge]{
        \begin{minipage}[t]{\linewidth}
            \centering
            \includegraphics[width=0.28\linewidth]{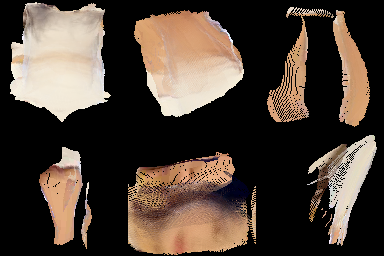}
            \includegraphics[width=0.28\linewidth]{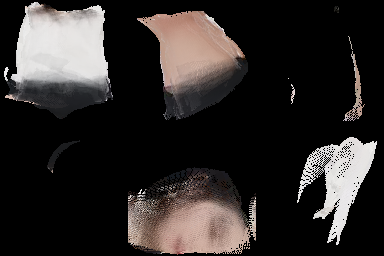}
            \includegraphics[width=0.28\linewidth]{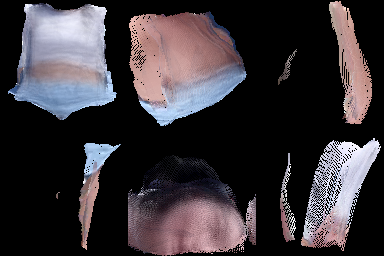}\label{fig:texture_compare_1}
        \end{minipage}
    }
    \subfigure[No Fine-tune]{
        \begin{minipage}[t]{\linewidth}
            \centering
            \includegraphics[width=0.28\linewidth]{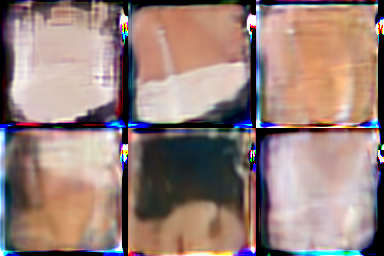}
            \includegraphics[width=0.28\linewidth]{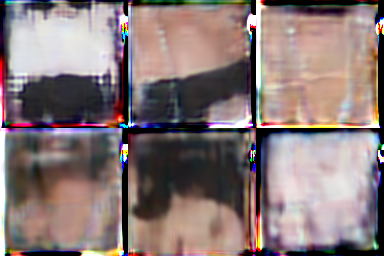}
            \includegraphics[width=0.28\linewidth]{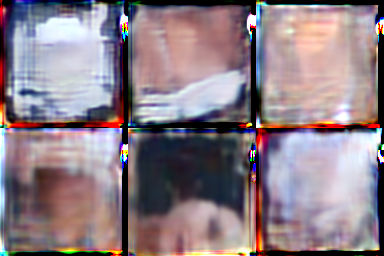}\label{fig:texture_compare_2}
        \end{minipage}
    }
    \subfigure[Fine-tune the Texture Map]{
        \begin{minipage}[t]{\linewidth}
            \centering
            \includegraphics[width=0.28\linewidth]{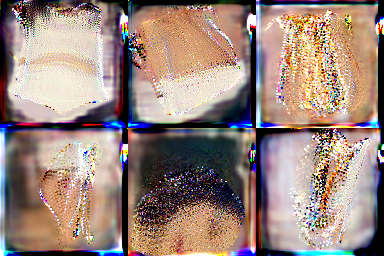}
            \includegraphics[width=0.28\linewidth]{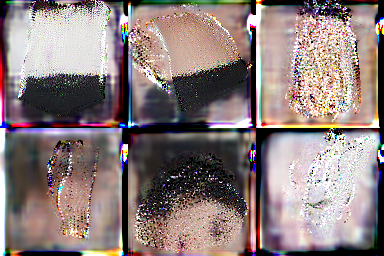}
            \includegraphics[width=0.28\linewidth]{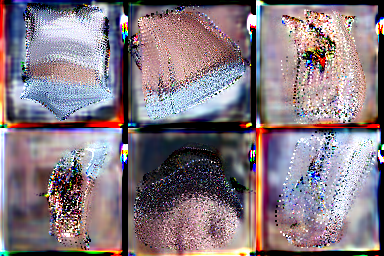}\label{fig:texture_compare_3}
        \end{minipage}
    }
    \subfigure[Fine-tune the Embedding]{
        \begin{minipage}[t]{\linewidth}
            \centering
            \includegraphics[width=0.28\linewidth]{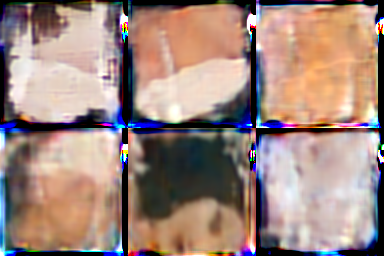}
            \includegraphics[width=0.28\linewidth]{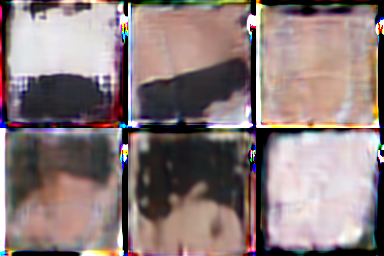}
            \includegraphics[width=0.28\linewidth]{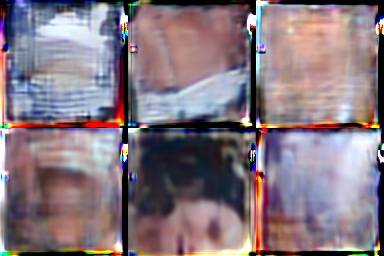}\label{fig:texture_compare_4}
        \end{minipage}
    }
    \caption{Texture maps of different texture map fine-tuning strategies. Corresponding video results can be found in the supplementary video.}
    \label{fig:texture_compare}
\end{figure*}

We show more examples of texture map in \Cref{fig:texture_compare} when different texture fine-tuning strategies are employed. Compare with the texture generator, direct merging source textures leaves a large proportion of the texture unfilled (Black part of \Cref{fig:texture_compare_1}). Directly fine-tuning the texture map adds noises to the texture. If it is rendered to the geometry, the synthesized human will be noisy and unnatural. Fine-tuning the embedding gives the most realistic texture with fewer artifacts and less noise. Comparison of generated videos with different texture maps can be found in the supplementary video file.

{\small
\bibliographystyle{ieee_fullname}
\bibliography{egbib}
\newpage
}

